\documentclass[sigconf]{acmart} 
\usepackage{epsfig}
\usepackage{graphicx}
\usepackage{amsmath}

\usepackage{amssymb}

\usepackage{booktabs}
\usepackage{multirow}
\usepackage{epstopdf}

\AtBeginDocument{%
  \providecommand\BibTeX{{%
    \normalfont B\kern-0.5em{\scshape i\kern-0.25em b}\kern-0.8em\TeX}}}

\setcopyright{acmcopyright}
\copyrightyear{2021}
\acmYear{2021}
\acmDOI{10.1145/1122445.1122456}

\acmConference[Woodstock '18]{Woodstock '18: ACM Symposium on Neural
  Gaze Detection}{June 03--05, 2018}{Woodstock, NY}
\acmBooktitle{Woodstock '18: ACM Symposium on Neural Gaze Detection,
  June 03--05, 2018, Woodstock, NY}
\acmPrice{15.00}
\acmISBN{978-1-4503-XXXX-X/18/06}

\usepackage{tabularx}
\begin{document}
	\newcolumntype{L}[1]{>{\raggedright\arraybackslash}p{#1}}
	\newcolumntype{C}[1]{>{\centering\arraybackslash}p{#1}}
	\newcolumntype{R}[1]{>{\raggedleft\arraybackslash}p{#1}}
\begin{sloppypar}
\title{Multi-View Correlation Distillation for Incremental Object Detection}

%
\author{Dongbao Yang}
\affiliation{%
 \institution{Institute of Information Engineering, Chinese Academy of Sciences}
 \institution{School of Cyber Security, University of Chinese Academy of Sciences}
 \city{Beijing}
 \country{China}}
\email{yangdongbao@iie.ac.cn}

\author{Yu Zhou}
\authornote{Corresponding author}
\affiliation{%
 \institution{Institute of Information Engineering, Chinese Academy of Sciences}
 \city{Beijing}
 \country{China}}
\email{zhouyu@iie.ac.cn}

\author{Weiping Wang}
\affiliation{%
 \institution{Institute of Information Engineering, Chinese Academy of Sciences}
 \city{Beijing}
 \country{China}}
\email{wangweiping@iie.ac.cn}

%
%
%
%
%
%



\renewcommand{\shortauthors}{Yang et al.}
\begin{abstract}
In real applications, new object classes often emerge after the detection model has been trained on a prepared dataset with fixed classes. Due to the storage burden and the privacy of old data, sometimes it is impractical to train the model from scratch with both old and new data. Fine-tuning the old model with only new data will lead to a well-known phenomenon of catastrophic forgetting, which severely degrades the performance of modern object detectors.
In this paper, we propose a novel \textbf{M}ulti-\textbf{V}iew \textbf{C}orrelation \textbf{D}istillation (MVCD) based incremental object detection method, which explores the correlations in the feature space of the two-stage object detector (Faster R-CNN). To better transfer the knowledge learned from the old classes and maintain the ability to learn new classes, we design correlation distillation losses from channel-wise, point-wise and instance-wise views to regularize the learning of the incremental model. 
A new metric named \textit{Stability-Plasticity-mAP} is proposed to better evaluate both the stability for old classes and the plasticity for new classes in incremental object detection. The extensive experiments conducted on VOC2007 and COCO demonstrate that MVCD 
can effectively learn to detect objects of new classes and mitigate the problem of catastrophic forgetting.

\end{abstract}





\maketitle


\section{Introduction}
\begin{figure}[t]
	
	\begin{minipage}[b]{0.45\linewidth}
		\centerline{\includegraphics[width=4.00cm]{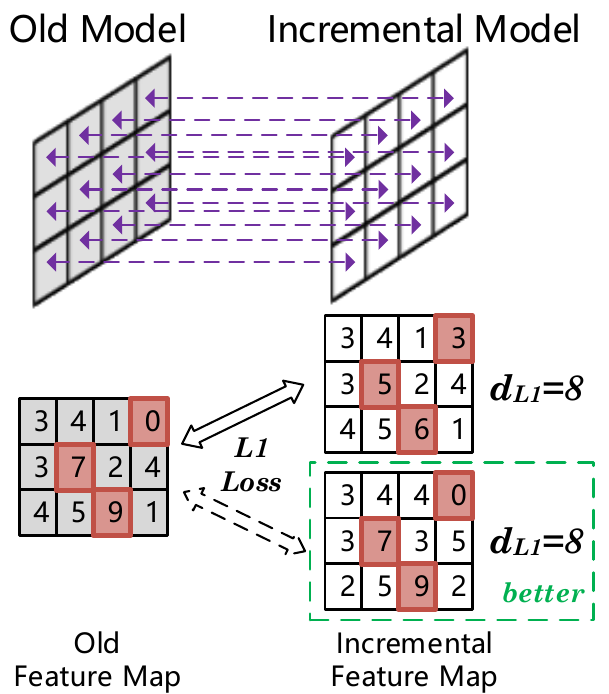}}
		\centerline{(a)}
	\end{minipage}
	\begin{minipage}[b]{0.52\linewidth}
		\centerline{\includegraphics[width=4.45cm]{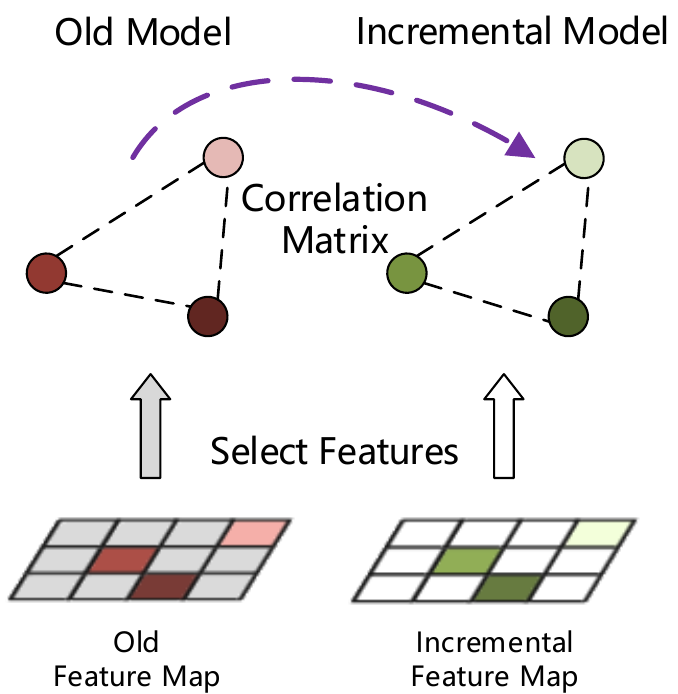}}
		\centerline{(b)}
	\end{minipage}
	
	\caption{Illustrations of (a) first-order distillation and (b) correlation distillation. Red and Green feature bins represent the important regions that should be focused on.}
	\label{fig:l1}
\end{figure}
Object detection is a basic computer vision task in many multimedia applications, such as autonomous driving, object tracking and so on. Modern object detection methods based on Convolution Neural Networks (CNNs) have achieved state-of-the-art results, which are usually trained on predefined datasets with a fixed number of classes. In many practical applications, new object classes often emerge after the detectors have been trained. Due to the privacy of data and limited storage of the devices, sometimes the old data can not be available for training the detectors from scratch. Even if the old data are available, this procedure will take a long training time. Fine-tuning is a commonly used method to transfer the pretrained model on new data. However, directly fine-tuning on new classes will severely decrease the performance on old classes~\cite{kirkpatrick2017overcoming}, which is known as catastrophic forgetting~\cite{french1999catastrophic}~\cite{goodfellow2013empirical}~\cite{mccloskey1989catastrophic}. Therefore, improving the ability of object detectors to learn new object classes continuously is necessary.

Recently, incremental learning has been paid more attention to classification, which aims to continuously learn to address new tasks from new data while preserving the learned knowledge from the old data. Based on the regularization methods to overcome catastrophic forgetting, the incremental learning methods can be divided into two categories~\cite{hou2019learning}: parameter-based~\cite{aljundi2018memory}~\cite{kirkpatrick2017overcoming}~\cite{zenke2017continual} and distillation-based~\cite{aljundi2017expert}~\cite{jung2018less}~\cite{li2017learning}~\cite{rannen2017encoder}~\cite{rebuffi2017icarl}. Due to the difficulty of designing a reasonable metric to evaluate the importance of all parameters, we follow the distillation-based regularization methods to preserve the learned knowledge from the old classes when adapting the old model on the data of new object classes.

\begin{figure}[t]
	\begin{center}
		\includegraphics[width=0.7\linewidth]{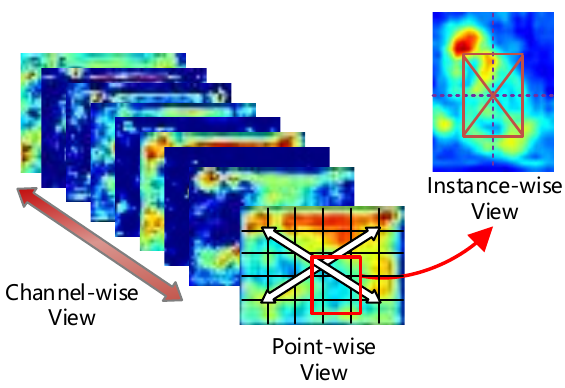}
	\end{center}
	\caption{Illustration of three views for correlation distillation.}
	\label{fig:mvc}
\end{figure}

Different from image classification, object detection involves distinguishing foreground from complex background and the precise localization of objects, which is more challenging for incremental learning. Existing incremental object detection methods~\cite{chen2019new}~\cite{hao2019take}~\cite{hao2019end}~\cite{li2019rilod}~\cite{shmelkov2017incremental}~\cite{zhang2020class} mainly adopt knowledge distillation to regularize the behavior of the incremental model to be similar to the old model for preserving the learned knowledge. The typical way is to minimize the distance between features or output logits of the old and incremental models.  
However, due to the inherent difference in the categories to be detected, we should preserve the stability to detect old classes and the plasticity to learn new classes. Directly enforcing the incremental model to imitate all activations in the feature maps of the old model (denoted as first-order distillation) makes the incremental model confusing about which knowledge is important and should be transferred. The sufficiently learned valuable knowledge may not be well preserved, instead some unimportant and misleading knowledge may be preserved. As shown in Figure~\ref{fig:l1}(a), the relative relations between the important activations are broken in the distillation procedure. 

In the perspective of the linguistic structuralism~\cite{matthews2001short}, the meaning of a sign depends on its relations with other signs within the system~\cite{park2019relational}. Analogously, the meaning of an activation value depends on its relation with other activation values within the feature map. An activation value will be meaningless without regard to its context activation values. 
Compared with the first-order distillation, the correlation distillation (second-order distillation) transfers a similarity correlation matrix 
as shown in Figure~\ref{fig:l1}(b), which explores intra-feature structural relations rather than individual activations and transfers a high-level representation of the activations in the feature maps. As studied in the representational similarity analysis in neuroscience~\cite{kriegeskorte2008representational}, the transformed feature contains more information than the original feature, which is the high-level abstractions of the activation behavior in the features of neural network~\cite{blanchard2019neurobiological}. 
For object detection, the relations within the activations in feature maps contain more information, such as image-level relations, foreground-background relations and intra-instance relations. Exploring and transferring the relative similarity of the discriminative patterns in the feature space can preserve the stability and plasticity for incremental object detection.

In this paper, we propose a novel multi-view correlation distillation based incremental object detection method (MVCD), which mainly focuses on the design of distillation losses in the feature space of the two-stage object detector Faster R-CNN~\cite{ren2015faster}. It is a dual network including the old model and the incremental model, which cooperate for transferring the old model trained on old classes to incrementally detect new classes. To trade off between the stability to preserve the learned knowledge from the old data and the plasticity to learn new knowledge from new classes, we design the correlation distillation losses from three views in the feature space of the object detector, which consists of the channel-wise, point-wise and instance-wise views as shown in Figure~\ref{fig:mvc}. Here, the three views for the feature maps can be seen as three abstractions of activation behaviors obtained from the stimuli from the different parts of the input. The channel-wise view explores the correlation among the feature map channels in the image-level feature. The point-wise view explores the knowledge among the discriminative regions corresponding to the foreground and the background. The instance-wise view explores the correlation among the intra-instance patches to preserve the discriminability of features for detecting the old classes.


The contributions of our work are as follows:
\begin{itemize}
	\setlength{\itemsep}{0pt}
	\setlength{\parsep}{0pt}
	\setlength{\parskip}{0pt}
	\item We propose a novel incremental object detection method, which is the first attempt to explore the multi-view correlations (second-order distillation) in the feature space of the two-stage object detector.
	\item To get a good trade-off between the stability and the plasticity of the incremental model, we design correlation distillation losses from three views for regularizing the learning in feature space, which transfers the learned channel-wise, point-wise and instance-wise correlations to the incremental model.  
	\item A new metric called \textit{Stability-Plasticity-mAP} (SPmAP) is proposed to quantize \textit{Stability} and \textit{Plasticity}, which is integrated with the original mAP to measure the performance of incremental object detector comprehensively.
	\item Extensive experiments are conducted on VOC2007~\cite{everingham2010pascal} and COCO~\cite{lin2014microsoft}. The results demonstrate the effectiveness of the proposed method to learn to detect new classes continuously, and it also achieves promising results compared with previous methods. 
\end{itemize}

\section{Related Work}
Incremental learning aims to develop machine learning systems to 
continuously deal with streams of new data while preserving the learned knowledge from the old data. 
The main challenge is to mitigate catastrophic forgetting and find a good trade-off between the stability and the plasticity of the incremental model. 

According to the optimization directions to preserve the learned knowledge, existing works can be divided into two categories~\cite{hou2019learning}: parameter-based and distillation-based. The parameter-based methods aim to preserve important parameters and penalize the changes of these parameters, such as EWC~\cite{kirkpatrick2017overcoming} and MAS~\cite{aljundi2018memory}. However, designing a metric to evaluate the importance of all parameters is also a tough task. Therefore, we mainly focus on distillation-based methods in our work.

\textbf{Distillation-based Incremental Learning:} Knowledge distillation is a commonly used technique to transfer knowledge from one network to another network. Hinton et al.~\cite{hinton2015distilling} propose to transfer the knowledge from a large network to a small network using distillation by encouraging the responses of these two networks to be similar. 
For incremental learning, distillation-based methods utilize the learned knowledge from the old model to guide the learning of the new model by minimizing the distillation losses. LwF~\cite{li2017learning} utilizes a modified cross-entropy loss to preserve original knowledge with only examples from the new task. iCaRL~\cite{rebuffi2017icarl} combines representation learning and knowledge distillation for jointly learning feature representation and classifiers, and a small set of exemplars is selected to perform nearest-mean-of-exemplars classification. Rannen et al.~\cite{rannen2017encoder} propose an auto-encoder based method to retain the knowledge from old tasks, which prevents the reconstructions of the features from changing and leaves space for the features to adjust. Sun et al.~\cite{sun2018active}~\cite{sun2018lifelong} propose to maintain a lifelong dictionary, which is used to transfer knowledge to learn each new metric learning task.

Recently, several novel knowledge distillation methods have explored the relationships between samples or instances to transfer the knowledge from teacher model to student model~\cite{li2020local}~\cite{liu2019knowledge}~\cite{park2019relational}. Liu et al.~\cite{liu2019knowledge} construct the instance relationship matrix. Park et al.~\cite{park2019relational} propose the distance-wise and angle-wise distillation losses to minimize the difference in relations. Li et al.~\cite{li2020local} propose to explore the local correlations to transfer the knowledge. Inspired by these methods, we believe that transferring the correlations in feature space for incremental learning may not only preserve the learned knowledge of the old model but also maintain the scalability to learn new knowledge, which can get a balance between stability and plasticity of the incremental model.   

\textbf{Incremental Object Detection:} The first incremental object detector~\cite{shmelkov2017incremental} is based on Fast R-CNN~\cite{girshick2015fast}. It uses EdgeBoxes~\cite{zitnick2014edge} and MCG~\cite{arbelaez2014multiscale} to precompute proposals, and knowledge distillation is used to regularize the outputs of the final classification and regression layers in order to preserve the performance on old classes. 
Recently, several end-to-end incremental object detection methods~\cite{chen2019new}~\cite{hao2019take}~\cite{hao2019end}~\cite{li2019rilod} are proposed. Chen et al.~\cite{chen2019new} propose to use L2 loss to minimize the difference between the feature maps of the old and the incremental models, which is referred to hint loss. 
Hao et al.~\cite{hao2019take} introduce a hierarchical large-scale retail object detection dataset called TGFS and presents a class-incremental object detector that utilizes an exemplar set with a fixed size of old data for training. Hao et al.~\cite{hao2019end} use a frozen duplication of RPN to preserve the knowledge gained from the old classes, and a feature-changing loss (L2 Loss) is proposed to reduce the difference of the feature maps between the old and new classes. Li et al.~\cite{li2019rilod} extract three types of knowledge from the original model, which is based on RetinaNet~\cite{lin2017focal}, and it uses smooth L1 loss to minimize the feature difference. A dual distillation training function is proposed in~\cite{zhang2020class} that pre-trains a separate model only for the new classes, such that a student model can learn from two teacher models simultaneously. In addition, several novel works on incremental few-shot object detection are proposed~\cite{perez2020incremental}~\cite{yang2020context}. 
However, the few-shot setting is more challenging than the many-shot setting in incremental object detection, and the problem of incremental object detection on the many-shot setting has not been well resolved. In our work, we mainly focus on general incremental learning for object detection. 

The typical way of the above-mentioned incremental object detection methods to preserve the learned knowledge is to imitate the important activations of the original model by minimizing the first-order distillation losses. However, it is hard for the incremental model to fully understand the transferred knowledge due to the inherent difference in the categories to be detected. Different from these methods, we explore the important correlations in the feature space of the object detector and only transfer the correlations instead of the values in the feature maps, which can preserve the relative relations within the important learned knowledge and maintain the capability to learn to detect new classes.

\section{Method}

\begin{figure}[t]
	\begin{center}
		\includegraphics[width=1.0\linewidth]{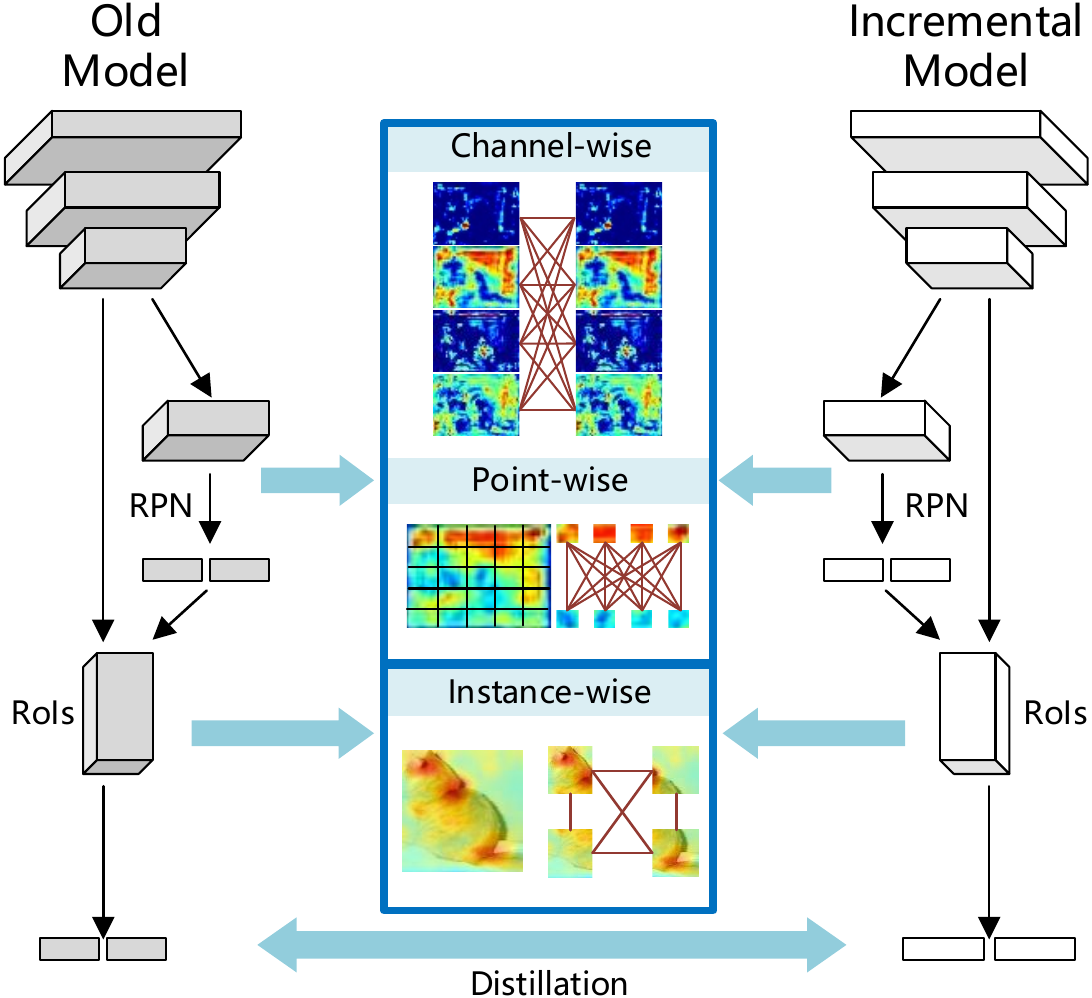}
	\end{center}
	\caption{The whole framework of the proposed method. The old model is a frozen copy of the original model trained on the old data, and the incremental model is an adapted model on new classes. Grey and white indicate the frozen and updated parameters respectively.}
	\label{fig:framework}
\end{figure}

\subsection{Overview}
The proposed multi-view correlation distillation mechanism for incremental object detection is shown in Figure~\ref{fig:framework}, which is a dual network. A frozen copy of the old model trained on the old data provides the learned knowledge of the old classes, such as the activations on feature maps, detection results and the logits from the output layers. The incremental model is adapted to detect both old and new classes on new data with the annotations of new classes as well as the learned knowledge from the old model. Samples of the new data are input into both the old model and the incremental model, then the detection results of the old model are integrated with the ground-truth of new classes to guide the learning of the incremental model. We also use the commonly used distillation loss (L1 loss) on the final output layers (classification layer and regression layer) to penalize the difference between the logits from the old model and the incremental model. In addition to these techniques, in our work, we mainly focus on the distillation in the feature space of the object detector to better preserve the learned knowledge. Different from image classification, the feature space in object detectors can be divided into image-level features and instance-level features, so we elaborately design the distillation losses for both of them to maintain the important knowledge.

\subsection{Multi-view Correlation Distillation}

The typical way to preserve the learned knowledge from the old model in feature space is to minimize the distance such as L1 loss between the activations of the old model and the incremental model. However, it is difficult for the incremental model to fully understand the transferred knowledge in feature space, which may result in the preservation of unimportant information instead of the useful knowledge for minimizing the overall loss. Meanwhile, this constraint may also restrict the plasticity of the incremental model for learning new classes. Incremental object detection aims to not only preserve the learned knowledge but also maintain the scalability for learning new classes. 
Therefore, we design a novel correlation distillation mechanism, which explores and transfers the important correlations from channel-wise, point-wise and instance-wise views in the feature space of the old object detector. The channel-wise view explores the correlation between the important feature maps in the image-level feature. The point-wise view explores the correlation between the discriminative background and foreground regions. The instance-wise view explores the correlation between the intra-instance patches, which aims to preserve the discriminability of the features for precisely detecting the old classes. The total loss function is defined as:
\begin{equation}
\label{Eq:total}
\begin{aligned}
\mathcal{L}=\mathcal{L}_{frcnn}+\mathcal{D}_{out}+\lambda(\mathcal{D}_{cc}+\mathcal{D}_{pc}+\mathcal{D}_{ic})
\end{aligned}
\end{equation}
where $\mathcal{L}_{frcnn}$ is the standard loss function in Faster R-CNN, and $\mathcal{D}_{out}$ is the commonly used distillation loss on the final classification and regression layers, and here we use L1 loss. $\mathcal{D}_{cc}$, $\mathcal{D}_{pc}$ and $\mathcal{D}_{ic}$ are the proposed channel-wise, point-wise and instance-wise correlation distillation losses. We set $\lambda=1$ in our experiments.

\subsubsection{Channel-wise Correlation Distillation}
\label{sec:cc}

The convolution kernels are responsible for extracting different patterns, so the channel-wise importances are different for each sample, which is the indispensable knowledge to transfer. However, the correlations between the feature distribution along channels are seldom considered in previous first-order-distillation-based methods. Intuitively, to preserve the plasticity of the incremental object detector, only the important channel-wise activations learned on the old data should be transferred to the incremental model and the rest unimportant channels can be left for learning new classes. Due to the disadvantage of first-order distillation, we propose a channel-wise correlation distillation loss. It constrains the specific inter-channel relations for different samples, and the consistent correlations of feature distribution along channels between the old and incremental model are preserved. It is achieved by distilling the correlations within the important channels of each image rather than restricting the overall activation values to be similar.    

\begin{figure}[t]
	\begin{center}
		\includegraphics[width=1.0\linewidth]{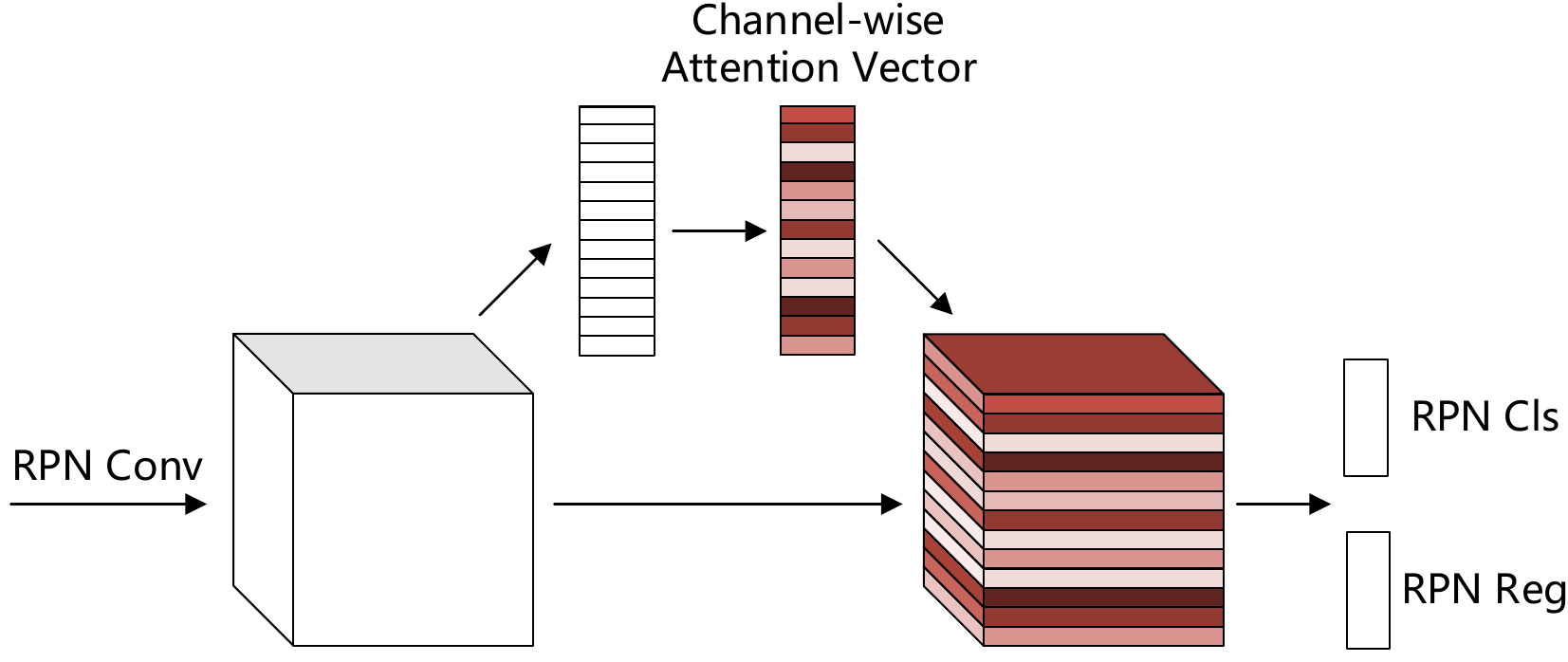}
	\end{center}
	\caption{SE block in RPN.}
	\label{fig:se}
\end{figure}
The squeeze-and-excitation module (SE module)~\cite{hu2018squeeze}, as a widely used channel attention module, can generate channel-wise weights for each image. SE module consists of squeeze and excitation operations. The original feature maps are aggregated across spatial dimensions, and a channel descriptor is obtained through a squeeze operation. Then, the sample-specific activations are learned for each channel (channel-wise attention vectors $\textbf{v}$) 
through an excitation operation. The original feature maps are then reweighted to generate the output of the SE block. 
To measure the importance of channels in image-level features, an SE module is added after the convolution layer in RPN which has higher-level features for discriminating foreground and background. The structure is shown in Figure~\ref{fig:se}. 

After the channel-wise attention vector $\textbf{v} = \left\{ v_1,v_2,...,v_C\right\}$ from the old model is obtained for each image, we normalize the vector to $[0,1]$. The important channels $F^{cc}\in \mathbb{R}^{N^{cc}\times W \times H}$ are classified by a threshold (0.5), where $N^{cc}$ is the number of important channels. 
%
The loss can be written as:

\begin{equation}
\begin{aligned}
S^{cc}(i,j)=\psi(F^{cc}(i),F^{cc}(j))&,\quad
S^{cc'}(i,j)=\psi(F^{cc'}(i),F^{cc'}(j))\\
%
\label{Eq:cc}
\mathcal{D}_{cc}=\frac{1}{N^{cc}\times N^{cc}}&\sum_{i=1}^{N^{cc}}\sum_{j=1}^{N^{cc}}|S^{cc}(i,j)-S^{cc'}(i,j)|
\end{aligned}
\end{equation}

where $S^{cc}$ is the channel-wise correlation matrix and $F$ and $F^{'}$ represent the features of the incremental model and the old model respectively. $i$ and $j$ are the indexes of the channels. $\psi(\cdot,\cdot)$ is cosine similarity between two vectors. The channel-wise correlation matrices of the old model and the incremental model both use the indexes of important channels obtained from the old model. 


The channel-wise correlations represent the relative distribution of specific patterns learned in different channels, which is the abstractions of channel-level activation behaviors. The channel-wise correlation distillation only transfers correlations between the important channels of old classes and removes redundant channel-wise information, which leaves room for learning new classes.

\subsubsection{Point-wise Correlation Distillation}
\label{sec:pc}
RPN is used to discriminate the object-like region proposals and background region proposals. The discriminative points in the activation-based spatial attention map of the feature in RPN correspond to the obvious regions of the background or foreground. Extracting the feature vectors of these points and only transferring the correlation between these point-wise feature vectors can preserve the knowledge of obvious foreground and background learned on the old data. The rest points are left to be optimized on new data. 

The obvious regions are the points with high or low responses on the activation-based spatial attention map, which can be obtained by thresholding the attention map. We use $F_{att}=\sum_{i=1}^{C}{|F_i|}$ to get the attention map, where $C$ is the number of channels. Then, the activation-based spatial attention map is further normalized for selecting the point-wise feature vectors with high or low responses. Here, we use two thresholds $\theta_{high}$ and $\theta_{low}$ to select these point-wise feature vectors of discriminative regions. The correlation matrix is constructed between them, which also uses the cosine similarity to describe the correlation.

In the dual network, the activation-based spatial attention map is obtained from the image-level feature of the old model ($F^{'}$), then the attention map is used to select points with high 
and low 
responses. 
After the indexes of these points are obtained $P^{high}=\left\{ (x_1,y_1),(x_2,y_2)...,(x_{N^h},y_{N^h})\right\}$ and $P^{low}=\left\{(x_1,y_1),(x_2,y_2)...,(x_{N^l},y_{N^l})\right\}$, where $N^h$ and $N^l$ are the numbers of the points with high and low responses respectively, we extract the point-wise feature vectors from the features of the old model and the incremental model ($F^{'}$ and $F$) respectively. The point-wise correlation matrices are also calculated for these two models respectively, and the distance between these two matrices is minimized. The loss is written as Equation~\ref{Eq:pc}. 


\begin{equation}
\begin{aligned}
S^{pc}(i,j)\!=\!\psi(F^{high}(i),F^{low}(j)),&\quad
S^{pc'}(i,j)\!=\!\psi(F^{high'}(i),F^{low'}(j))\\
\label{Eq:pc}
\mathcal{D}_{pc}=&||S^{pc}-S^{pc'}||_F^2
%
\end{aligned}
\end{equation}

where $F^{high}\in \mathbb{R}^{N^h \times C}$ and $F^{low}\in \mathbb{R}^{N^l \times C}$ are the extracted point-wise feature vectors corresponding to the points with high and low responses from the incremental model respectively. Similarly, $F^{high'}\in \mathbb{R}^{N^h \times C}$ and $F^{low'}\in \mathbb{R}^{N^l \times C}$ represent the corresponding feature vectors from the old model.

The point-wise correlations represent the abstractions of the activation behaviors about the relative responses between foreground and background. The point-wise correlation distillation can preserve the consistent correlations between the feature distributions of obvious foreground and background, and the indistinct regions can be left to learn new classes.     

\subsubsection{Instance-wise Correlation Distillation}
\label{sec:ic}
Inspired from~\cite{li2020local}, the local features and their correlation in each instance also contain many details and discriminative patterns. 
The old model can generate more discriminative local features with sufficient old data, while the incremental model is hard to achieve that due to the lack of old data. 
The old model trained on old classes can make right predictions for different categories of objects with similar appearances based on the discriminative local regions of each instance. Therefore, learning the local knowledge of each instance from the old model is an important way to maintain the stability to detect the old classes. The intra-instance correlation is not considered when simply imitating the global activations of instance-level features, which will degrade the discriminability for detecting old classes due to the loss of distinctive local patterns.   


To transfer the correlation of the local regions for each instance from the old model to the incremental model, we compute the correlation matrix of the local regions for each instance-level feature, which is the pooled feature after the RoI-Pooling and convolution layers in the detection head. The pooled feature of each instance $F^l \in \mathbb{R}^{pc \times ph \times pw}$ ($pc=2048$, $ph=4$ and $pw=4$) are divided into $k \times k$ ($k=2$) patches, and each patch has a shape of $pc \times \frac{ph}{k} \times \frac{pw}{k}$ ($2048\times2\times2$). The instance-wise correlation distillation loss is defined as:


\begin{equation}
\begin{aligned}
S^{ic}(i,j)=\psi(F^l(i),F^l(j))&,\quad
S^{ic'}(i,j)=\psi(F^{l'}(i),F^{l'}(j))\\
\label{Eq:ic}
\mathcal{D}_{ic}=\frac{1}{k\times k}\sum_{i=1}^{k}&\sum_{j=1}^{k}|S^{ic}(i,j)-S^{ic'}(i,j)|
\end{aligned}
\end{equation}
where $F^l(\cdot)$ and $F^{l'}(\cdot)$ are the vectorized features of the local patches from the incremental model and the old model respectively.

The instance-wise correlations represent the abstractions of activation behaviors within an instance, which explores the correlations between the local parts of an instance. The instance-wise correlation distillation can transfer the relative relationship between the response values of class-specific local features learned from the old model.

\subsection{Stability-Plasticity-mAP}
For incremental object detection, due to the different numbers of old and new classes and the different difficulties of learning each class, mAP is not very suitable for measuring the performance of the incremental model on handling the stability-plasticity dilemma. Therefore, to quantize \textit{Stability} and \textit{Plasticity}, we propose a new metric called \textit{Stability-Plasticity-mAP} (SPmAP), as written in Equation~\ref{Eq:metric}. Because incremental object detection aims to reach the performance of the model trained on all data with only the data of new classes, we use the model train on all classes as the up-bound model to measure stability and plasticity. 
\textit{Stability} is the average difference of precisions on old classes, and \textit{Plasticity} is the average difference of precisions on new classes. 
We also integrate the overall mAP difference ($mAP_{dif}$) representing the overall performance of all classes into the metric to measure the performance comprehensively.
\begin{equation}
\begin{aligned}
SPmAP=((Stability&+Plasticity)/2+mAP_{dif})/2 \\
Stability=\frac{1}{N^{o}}&\sum\nolimits_{i=1}^{N^{o}}(UP(i)-INC(i)) \\
Plasticity=\frac{1}{N^{n}}&\sum\nolimits_{i=N^{o}+1}^{N}(UP(i)-INC(i)) \\
mAP_{dif}=\frac{1}{N}&\sum\nolimits_{i=1}^{N}\!(UP(i)\!-\!INC(i))	
\end{aligned}
\label{Eq:metric}
\end{equation}
where $N$ is the number of all classes. $N^{o}$ and $N^{n}$ represent the numbers of old and new classes respectively. $UP$ and $INC$ are the average precisions of the up-bound model and the incremental model. 

\section{Experiments and Evaluation}


\begin{table*}

	\caption{Per-class average precision (\%) on VOC2007 test dataset. Comparisons are conducted under different settings when 1, 5, 10 classes are added at once. }
	\begin{center}	
		\resizebox{1.0\textwidth}{!}{
			\begin{tabular}{l|cccccccccccccccccccccc}
				\toprule
				
				\multicolumn{23}{c}{1}\\
				\midrule\specialrule{0.0em}{0pt}{0pt}
				Method &aero & bike & bird & boat & bottle & bus & car & cat & chair & cow & table & dog & horse & mbike & person & plant & sheep & sofa & \multicolumn{1}{c|}{train} & \multicolumn{1}{c|}{tv} &\multicolumn{1}{c|}{ mAP} & SPmAP\\
				\midrule\specialrule{0.0em}{0pt}{0pt}
				Old(1-19) &74.3&77.9 &	72.6 & 59.3 & 52.5 & 77.7 & 81.3 & 84.8 &	45.1 &	82.7 &	63.3 &	85.0 &	83.6 &	78.5 &	77.1 &	41.4 &	72.2 &	66.4 &	\multicolumn{1}{c|}{77.9} &\multicolumn{1}{c|}{-} &\multicolumn{1}{c|}{71.2}&- \\
				\midrule\specialrule{0.0em}{0pt}{0pt}
				Fine-tuning &31.2&	24.7 &	30.1 &	21.4 &	21.9 & 49.7 &	58.3 & 28.3 &	11.9 &	33.9 &	7.7 & 21.4 &	18.6 & 17.0 &	14.5 &	10.9 &	28.5 &	9.8 &\multicolumn{1}{c|}{25.4} &	\multicolumn{1}{c|}{59.8} &	\multicolumn{1}{c|}{26.2}&30.4 \\
				Shmelkov et al.~\cite{shmelkov2017incremental} & 65.0 & 72.2 &	69.4 &	53.1 &	50.4 &	71.8 &	78.5 &	81.9 &	46.8 &	79.3 &	57.3 &	82.0 &	81.9 &	74.4 & 76.2 & 39.3 &	70.0 &	62.3 &	\multicolumn{1}{c|}{60.3} &	\multicolumn{1}{c|}{60.3} &	\multicolumn{1}{c|}{66.6}&5.8\\
				Chen et al.~\cite{chen2019new}& 70.7  &76.8  & 72.5  &	57.8  &	51.0  &	75.3  &	81.4  &	84.7  &	46.7  &	80.0  &	61.4  &	82.4  &	83.3  &	75.8  & 76.4  &	40.5  &	73.1 & 63.9 &	\multicolumn{1}{c|}{69.8 } &	\multicolumn{1}{c|}{51.6 } &	\multicolumn{1}{c|}{68.8 }&6.3	\\
				Hao et al.~\cite{hao2019take}&67.5 &	73.4 &	66.9 & 51.8 &	49.4 &	70.8 &	77.8 &	82.6 &	45.9 &	81.2 &	60.1 &	80.4 &	82.2 & 74.9 &	76.5 & 38.0 &	70.7 &	64.7 &	\multicolumn{1}{c|}{60.6} &	\multicolumn{1}{c|}{60.5} &	\multicolumn{1}{c|}{66.8 }&5.7	\\
				Hao et al.~\cite{hao2019end}&69.5 & 76.5 &	69.9 &	57.1 &	49.8 & 74.3 &	79.3 &	79.7 &	46.9 &	82.5 &	61.4 &	81.5 &	82.0 &	75.1 &	76.7 &	39.3 &	73.8 &	64.5  &	\multicolumn{1}{c|}{66.2} &	\multicolumn{1}{c|}{58.6 } &	\multicolumn{1}{c|}{68.2 }	&5.0\\
				
				Li et al.~\cite{li2019rilod}&71.3 &	75.8 &	70.8 &	56.2 &	50.0 &	74.8 &	80.0 &	82.3 &	46.5 &	83.0 &	58.5 &	82.5 & 79.9 & 78.1 &	77.0 & 39.4 &	72.4 &	66.1 &	\multicolumn{1}{c|}{69.0} &	\multicolumn{1}{c|}{59.7} &	\multicolumn{1}{c|}{68.7 }&4.4\\
				
				Plain L1&70.6 &	77.1 &	70.7 &	58.4 &	50.4 &	75.2 &	80.5 &	83.0 &	46.5 &	83.0 &	61.9 &	82.7 &	81.4 &	74.9 &	76.9 &	40.1 &	72.3 &	65.4 &	\multicolumn{1}{c|}{67.6} 	&	\multicolumn{1}{c|}{59.6 }&	\multicolumn{1}{c|}{68.9} &4.2\\
				MVCD &71.2 & 76.7 &	71.3 &	60.1 &	51.2 &	76.7 &	80.2 &	83.5 &	47.4 &	82.4 &	62.5 &	83.2 &	83.2 &	75.9 &	77.2 &	41.6 &	72.0 &	66.6 &		\multicolumn{1}{c|}{70.7} &		\multicolumn{1}{c|}{60.6} &	\multicolumn{1}{c|}{\textbf{69.7}} &\textbf{3.4}\\
				\midrule\specialrule{0.0em}{0pt}{0pt}
				\multicolumn{23}{c}{5}\\

				\midrule\specialrule{0.0em}{0pt}{0pt}
				Method& aero & bike & bird & boat & bottle & bus & car & cat & chair & cow & table & dog & horse & mbike &\multicolumn{1}{c|}{person} & plant & sheep & sofa & train & \multicolumn{1}{c|}{tv} & \multicolumn{1}{c|}{mAP}&SPmAP\\
				\midrule\specialrule{0.0em}{0pt}{0pt}
				Old(1-15) &72.8&77.5 &	72.5 &	58.7 &55.2 &	74.8 &	83.7 &	85.6 &	47.2 &	76.3 &	64.2 &	82.9 &	82.7 &78.1 &\multicolumn{1}{c|}{77.2} &-&-&-&-&\multicolumn{1}{c|}{-}&\multicolumn{1}{c|}{	72.6} &-\\
				\midrule\specialrule{0.0em}{0pt}{0pt}
				Fine-tuning &42.2&	38.4 &	38.9 & 33.4 &	22.1 &	27.8 &	64.0 & 62.9 &	16.8 &	39.5 &	35.4 &	50.0 &	59.9 &	42.1 &	\multicolumn{1}{c|}{25.8} &33.9 &	53.2 &	53.4 &	51.8 &	\multicolumn{1}{c|}{62.1} &\multicolumn{1}{c|}{42.7} &26.6\\				
				Shmelkov et al.~\cite{shmelkov2017incremental} &57.9  & 74.0  &	65.9  &	39.7  &	47.4  &	46.4  &	78.2  & 77.7  &	44.8  &	68.9  &	59.2  &	76.0  &	80.8  &	73.2  &	\multicolumn{1}{c|}{74.7}  &	36.3  &	62.1  &	61.4  &	64.3 &	\multicolumn{1}{c|}{64.2}&	\multicolumn{1}{c|}{62.7 }&8.8\\
				Chen et al.~\cite{chen2019new}&65.0&75.8 &	67.0 &	46.1 &	51.8 &	54.5 &	80.5 &	79.2 &	46.0 &	72.1 & 62.8 &	74.4 &	81.3 &	75.3 &	\multicolumn{1}{c|}{74.9} &	32.2 &	60.6 &	55.0 &	54.5 &	\multicolumn{1}{c|}{56.5} &	\multicolumn{1}{c|}{63.3} &9.2\\
				Hao et al.~\cite{hao2019take}&56.5 &	74.6 &	67.0 &	39.7 &	47.3 &	53.4 &	77.9 &	79.4 &	44.6 &	68.6 &	56.9 &	75.7 &	80.4 &	75.5 &\multicolumn{1}{c|}{	74.9} &	36.5 &	60.8 &	59.9 &	65.2 &	\multicolumn{1}{c|}{64.4} &	\multicolumn{1}{c|}{63.0} &8.6\\
				Hao et al.~\cite{hao2019end}&61.4& 75.2 &	67.4 &	44.3 &	49.6 &	51.7 & 79.2 &	78.7 &	45.8 &	70.7 &	61.0 & 76.0 &	81.8 & 75.4 &	\multicolumn{1}{c|}{74.8} &	36.3 &	63.9 &	59.4 &	66.1 & \multicolumn{1}{c|}{63.6} &\multicolumn{1}{c|}{	64.1}&7.5 \\
				
				Li et al.~\cite{li2019rilod}&63.8& 76.5 &70.1 & 48.2 &	50.7 &	54.9 &	80.6 &	79.6 &	45.7 & 74.7 &	59.7 &	78.2 &	82.7 &	73.0 &	\multicolumn{1}{c|}{75.2} &	36.8 &	64.0 &	63.2 &	65.3 &	\multicolumn{1}{c|}{61.5} &	\multicolumn{1}{c|}{65.2 }&6.5\\		
				Plain L1&64.0 &	75.1 &	68.5 &	45.8 &	51.8 &	52.7 &	80.0 &	78.5 &	45.8 &	72.8 &	60.3 &	76.8 &	81.6 &	75.7 &	\multicolumn{1}{c|}{75.2} &	36.3 &	63.1 &	59.3 &	64.6 &	\multicolumn{1}{c|}{61.3} &	\multicolumn{1}{c|}{64.5} &7.4\\
				MVCD& 65.7 &  76.6 &	71.9 &	51.5 &	51.0 &	64.9 &	81.5 &	80.6 &	47.0 &	74.3 &	60.8 &	80.6 &	82.2 &	76.8 &		\multicolumn{1}{c|}{75.8} &	37.0 &	63.9 &	58.9 &	67.0 &	\multicolumn{1}{c|}{62.8} &	\multicolumn{1}{c|}{\textbf{66.5} }&\textbf{5.5} \\
				
				\midrule\specialrule{0.0em}{0pt}{0pt}
				\multicolumn{23}{c}{10}\\
				\midrule
				Method& aero & bike & bird & boat & bottle & bus & car & cat & chair & \multicolumn{1}{c|}{cow} & table & dog & horse & mbike & person & plant & sheep & sofa & train & \multicolumn{1}{c|}{tv} &\multicolumn{1}{c|}{ mAP}&SPmAP\\
				\midrule\specialrule{0.0em}{0pt}{0pt}
				Old(1-10) &
				89.4  &	89.6 &	89.4 &87.2 &76.7 &83.9 &89.5 &86.6 &78.8 &\multicolumn{1}{c|}{71.8} &-&-&-&-&-&-&-&-&-&\multicolumn{1}{c|}{-}&\multicolumn{1}{c|}{84.3} &-\\
				\midrule
				Fine-tuning& 46.9 &33.2 &37.9 &	33.5 &28.3 &	45.2 &	45.9 &	52.3 &	13.0 &	\multicolumn{1}{c|}{35.6} &	50.0 &	64.6 &75.0 & 67.3 & 73.5 & 32.8 &	60.2 &59.2 &	61.9 &	\multicolumn{1}{c|}{59.4}&	\multicolumn{1}{c|}{48.8 }&22.8\\
				Shmelkov et al.~\cite{shmelkov2017incremental}&67.7 &	63.3 &	64.0 &	46.1 &	51.6 &	72.9 &	79.0 &	72.8 & 37.1 &\multicolumn{1}{c|}{	67.7} &	52.3 &	77.7 &	78.2 &	76.3 &	75.4 &37.6 &	67.6 &	67.2 &	74.0 &	\multicolumn{1}{c|}{65.6} &\multicolumn{1}{c|}{	64.7}&6.9 \\
				Chen et al.~\cite{chen2019new}&69.9 & 64.4 & 66.5 &	51.2 &	54.3 &	76.0 &	79.6 &	74.7 &	38.8 &\multicolumn{1}{c|}{ 71.7} &	51.0 &	77.3 &	78.6 &	73.7 &	72.9 &	29.1 &	65.7 &	63.9 &	71.0 &	\multicolumn{1}{c|}{57.2} &	\multicolumn{1}{c|}{64.4} &7.2\\
				Hao et al.~\cite{hao2019take}&65.3 & 64.2 &	64.7 &	48.7 &	50.6 &	72.8 &	79.4 &	72.0 &	37.6 &	\multicolumn{1}{c|}{67.4} &	50.9 &	78.1 &	78.3 &	76.9 &	75.4 &	39.2 &	68.2 &	63.8 &	72.7 &	\multicolumn{1}{c|}{66.9} &	\multicolumn{1}{c|}{64.6}&6.9\\
				Hao et al.~\cite{hao2019end}&68.4 &	61.9 &	67.0 &	52.7 &	53.3 &	73.2 &	80.2 &	74.8 & 38.7 &\multicolumn{1}{c|}{	69.5} &	54.7 &	75.1 &	78.4 &	76.1 & 74.7 &	34.2 &	69.7 &	65.2 &	71.7 &	\multicolumn{1}{c|}{65.1} &	\multicolumn{1}{c|}{65.2} &6.3\\
				
				Li et al.~\cite{li2019rilod}&70.1 &	64.0 &	68.0 &	52.6 &	52.6 &	73.5 &	81.1 &	75.7 &	39.0 &\multicolumn{1}{c|}{	66.7} &	53.9 &	77.7 &	78.9 &	74.5 &	73.7 &	33.4 &	67.0 &	63.6 &	70.5 &	\multicolumn{1}{c|}{63.2} &\multicolumn{1}{c|}{	65.0} &6.6\\
				Plain L1&71.2 &	64.4 &	67.4 &  53.5 &	53.3 &	76.0 &	79.9 &	76.9 &	38.9 &	\multicolumn{1}{c|}{71.1} &	54.7 &	78.0 &	78.7 &	74.1 &	73.8 &	34.3 &	67.4 &	65.1 &	70.8 &	\multicolumn{1}{c|}{62.8} &	\multicolumn{1}{c|}{65.6}  &5.9\\
				MVCD& 72.1 &
				68.9 &
				68.2 &
				53.9 &
				54.2 &
				74.7 &
				81.5 &
				75.3 &
				40.0 &
				\multicolumn{1}{c|}{72.7} &
				55.9 &
				79.5 &
				79.4 &
				73.5 &
				72.5 &
				32.9 &
				68.4 &
				61.4 &
				73.0 &
				\multicolumn{1}{c|}{63.7} &
				\multicolumn{1}{c|}{\textbf{66.1}} &
				\textbf{5.5}\\
				\midrule\specialrule{0.0em}{0pt}{0pt}
				Up-bound(1-20)&72.4 &76.9&73.4 &59.2&	54.5 &79.1 &	81.9 &	86.3 &	47.4 &	82.4 &	63.7 &	84.9 &	83.0 &	80.2 &	77.2 &	42.6 &	75.1 &	64.4 &	77.7 &	\multicolumn{1}{c|}{69.0} &	\multicolumn{1}{c|}{71.6}&- \\
				\bottomrule
		\end{tabular}}
	\end{center}
	
	\label{table:all-detail}
\end{table*}

\begin{table}
	\caption{Average precision (\%) on COCO minival (first 5000 validation images). Comparisons are conducted when 40 classes are added at once.}
	\begin{center}
		\resizebox{0.9\linewidth}{!}{
			\begin{tabular}{L{3cm}|C{1.5cm}C{1.5cm}}
				\toprule
				Method & mAP &SPmAP\\
				\midrule\specialrule{0.0em}{0pt}{0pt}
				Old(1-40) & 	52.26 &-\\
				\midrule\specialrule{0.0em}{0pt}{0pt}
				Fine-tuning&17.85 &31.74\\
				Plain L1& 44.46&5.00\\
				MVCD &\textbf{44.62}&\bf4.59\\
				\midrule\specialrule{0.0em}{0pt}{0pt}
				Up-bound(1-80) & 49.21&-\\
				\bottomrule
		\end{tabular}}
	\end{center}
	
	\label{table:coco}
\end{table}

%

\begin{table*}
	\caption{ Average precision (\%) on VOC2007 test dataset when adding 5 or 10 new classes sequentially.}
	\begin{center}
		\resizebox{0.78\textwidth}{!}{
			\begin{tabular}{c|L{1.3cm}|c|c|c|c|c|c|c|c|c|c}
				\toprule
				\multirow{3}{*}{5}&\multicolumn{1}{l|}{Method} &\multicolumn{2}{c|}{+plant} & \multicolumn{2}{c|}{+sheep} & \multicolumn{2}{c|}{+sofa}&\multicolumn{2}{c|}{+train} & \multicolumn{2}{c}{+tv}  \\
				\cmidrule{2-12}\specialrule{0.0em}{0pt}{0pt}
				&Plain L1&\multicolumn{2}{c|}{66.78}	&\multicolumn{2}{c|}{62.38}	&\multicolumn{2}{c|}{61.09}	&\multicolumn{2}{c|}{56.10}	&\multicolumn{2}{c}{51.39}\\
				&MVCD&\multicolumn{2}{c|}{\bf 68.19} &\multicolumn{2}{c|}{\bf 65.63}&	\multicolumn{2}{c|}{\bf 63.31} &	\multicolumn{2}{c|}{\bf 56.72} &	\multicolumn{2}{c}{\bf 51.89} \\ 
				\midrule\specialrule{0.0em}{0pt}{0pt}%
				\multirow{6}{*}{10}&\multicolumn{1}{l|}{Method}&+table & +dog & +horse & +mbike & +person & +plant & +sheep & +sofa & +train & +tv \\
				\cmidrule{2-12}\specialrule{0.0em}{0pt}{0pt}
				&Plain L1&58.34	&55.69	&50.14&	44.04&	39.59	&33.27&	31.89&	30.15&	25.46&	26.88\\			
				&MVCD &
				\bf 59.91 &\bf	58.51	 &\bf55.32 &	\bf50.18 &	\bf46.85 &	\bf40.05	 &\bf34.07 &	\bf32.53 &	\bf27.77 &	\bf27.13\\

				\cmidrule{2-12}\specialrule{0.0em}{0pt}{0pt}
				&\multicolumn{1}{l|}{Method} &\multicolumn{2}{c|}{+table \& dog} & \multicolumn{2}{c|}{+horse \& mbike} & \multicolumn{2}{c|}{+person \& plant}&\multicolumn{2}{c|}{+ sheep \& sofa} & \multicolumn{2}{c}{+train \& tv} \\
				\cmidrule{2-12}\specialrule{0.0em}{0pt}{0pt}
				&Plain L1&\multicolumn{2}{c|}{58.87}	&	\multicolumn{2}{c|}{53.42}	&	\multicolumn{2}{c|}{46.28}	&	\multicolumn{2}{c|}{46.50}	&	\multicolumn{2}{c}{42.67}\\
				&MVCD&\multicolumn{2}{c|}{\bf60.76}&		\multicolumn{2}{c|}{\bf59.20}&		\multicolumn{2}{c|}{\bf51.68}&		\multicolumn{2}{c|}{\bf51.78}&		\multicolumn{2}{c}{\bf48.23}\\
				
				\bottomrule
		\end{tabular}}
	\end{center}
	
	\label{table:10-sqe}
\end{table*}

\begin{table}
	\renewcommand\arraystretch{0.9}
	\caption{Results on VOC2007 and COCO, when four groups are added sequentially.}
	\begin{center}
		\resizebox{0.9\linewidth}{!}{
			\begin{tabular}{l|L{1.3cm}|cccc|c}
				\toprule
				&Method &A& B &C& D&mAP\\
				\cmidrule{1-7}\specialrule{0.0em}{0pt}{0pt}
				\multirow{8}{*}{VOC}
				&\multirow{4}{*}{Plain L1}  
				&  48.75    &	-    &	-    &	-   &  48.75   \\
				&& 38.96    &  60.55 &	-    &	-   &  49.75\\
				&& 22.01    &  27.57 & 55.09 &	-       &34.89\\
				&& 6.03     &  11.69 & 33.06 &  35.81	&21.65\\
				\cmidrule{2-7}\specialrule{0.0em}{0pt}{0pt}
				&\multirow{4}{*}{MVCD}
				& 48.75     &	-    &	-    &-	    &48.75 \\	 
				&&	\bf44.62   &    58.38     &	-    &	-   &\bf51.50 \\
				&&  \bf30.59    &    \bf33.08     &   \bf55.62    &	   -&\bf39.76\\
				&&  \bf14.25   &    \bf18.34     & \bf41.72      & 	\bf36.32     &\bf27.66\\
				\midrule\specialrule{0.0em}{0pt}{0pt}
				\multirow{8}{*}{COCO}
				&\multirow{4}{*}{Plain L1}  
				&  57.29    &	-    &	-    &	-   &57.29\\
				&& 45.24    &  19.48  	 &	-    &	-   &32.36\\
				&& 30.22    &  14.93     & 23.22      &	-   &22.79\\
				&& 19.14    &  10.58  	 & 14.02  	 &	28.49    &18.06\\			
				\cmidrule{2-7}\specialrule{0.0em}{0pt}{0pt}
				&\multirow{4}{*}{MVCD}
				& 57.29   &	- &	- &-	&57.29\\	 
				&&\bf48.48	 &\bf20.74    &	- &	-   &\textbf{34.61}\\
				&& \bf34.75  &  \bf15.45  &\bf24.29   &	-   &\textbf{24.83}\\
				&& \bf21.50  & \bf10.62   & \bf16.28  & \bf29.26    &\textbf{19.41}\\
				\bottomrule
		\end{tabular}}
	\end{center}
	
	\label{table:4-group}
\end{table}

\begin{table*}
	\caption{Ablation Study}
	\begin{center}
		\resizebox{0.70\linewidth}{!}{
			\begin{tabular}{ccc|cccccc}
				\toprule
				\multicolumn{3}{c|}{Correlation Distillation}&\multicolumn{2}{c|}{1} & \multicolumn{2}{c|}{5} &\multicolumn{2}{c}{10}\\
				\midrule\specialrule{0.0em}{0pt}{0pt}
				\multirow{1}{*}{$\mathcal{D}_{cc}$}&\multirow{1}{*}{$\mathcal{D}_{pc}$}&\multirow{1}{*}{$\mathcal{D}_{ic}$}
				&mAP&\multicolumn{1}{c|}{SPmAP}&mAP&\multicolumn{1}{c|}{SPmAP}&mAP&\multicolumn{1}{c}{SPmAP}\\
				\midrule
				&&&66.50&\multicolumn{1}{c|}{5.39}&62.90&\multicolumn{1}{c|}{8.50}&64.31&\multicolumn{1}{c}{7.24}\\
				\midrule\specialrule{0.0em}{0pt}{0pt}
				$\checkmark$&&&67.52(+1.02)&\multicolumn{1}{c|}{5.12}&63.28(+0.38) &\multicolumn{1}{c|}{8.19}&65.45($+1.14$)&\multicolumn{1}{c}{6.10}\\
				&$\checkmark$&&68.44(+1.94)&\multicolumn{1}{c|}{4.26} &65.28(+2.38)&\multicolumn{1}{c|}{6.37}&65.53($+1.22$)&\multicolumn{1}{c}{6.02}\\
				&&$\checkmark$&68.48(+1.98)&\multicolumn{1}{c|}{4.03}&65.49(+2.59)&\multicolumn{1}{c|}{6.29}&65.72($+1.41$)&\multicolumn{1}{c}{5.83}\\
				
				&$\checkmark$&$\checkmark$&69.47(+2.97)&\multicolumn{1}{c|}{3.52}&66.42(+3.52)&\multicolumn{1}{c|}{\bf 5.42}&65.77($+1.46$)&\multicolumn{1}{c}{5.78}\\
				$\checkmark$&&$\checkmark$&69.32(+2.82)&\multicolumn{1}{c|}{3.68}&66.48(+3.58)&\multicolumn{1}{c|}{5.52}&65.97($+1.66$)&5.58\\
				$\checkmark$&$\checkmark$&&68.69(+2.19)&\multicolumn{1}{c|}{4.33}&65.52(+2.62)&\multicolumn{1}{c|}{6.25}&65.70($+1.39$)&\multicolumn{1}{c}{5.85}\\
				\midrule\specialrule{0.0em}{0pt}{0pt}
				$\checkmark$&$\checkmark$&$\checkmark$&\textbf{69.70(+3.20)}&\multicolumn{1}{c|}{\bf 3.41}&\textbf{66.53(+3.63)}&\multicolumn{1}{c|}{\underline{ 5.49}}&\textbf{66.08(+1.77)}&\multicolumn{1}{c}{\bf 5.47}\\
				\bottomrule
		\end{tabular}}
	\end{center}
	
	\label{table:ablation}
\end{table*}

%
%
%

\begin{table}
	\caption{The results of alternative choices.}
	\begin{center}
		\resizebox{1\linewidth}{!}{
			\begin{tabular}{ll|cccccc}
				\toprule
				
				\multicolumn{2}{l|}{\multirow{2}{*}{Method}}&\multicolumn{2}{c|}{1}&\multicolumn{2}{c|}{5}&\multicolumn{2}{c}{10}\\
				\specialrule{0.0em}{0pt}{0pt}
				\cmidrule{3-8}\specialrule{0.0em}{-1pt}{-1pt}
				&&mAP&\multicolumn{1}{c|}{SPmAP}&mAP&\multicolumn{1}{c|}{SPmAP}&mAP&\multicolumn{1}{c}{SPmAP}\\
				\midrule\specialrule{0.0em}{0pt}{0pt}
				\multirow{1}{*}{$\mathcal{D}_{cc}$}
				&$L1$&70.00&\multicolumn{1}{c|}{3.88}&66.37&\multicolumn{1}{c|}{5.86}&65.21&\multicolumn{1}{c}{6.34}\\
				\midrule\specialrule{0.0em}{0pt}{0pt}
				\multirow{2}{*}{$\mathcal{D}_{pc}$}
				&$L1$&69.89 &\multicolumn{1}{c|}{4.40} &66.05 &\multicolumn{1}{c|}{ 6.31}& 64.58&\multicolumn{1}{c}{ 6.97}\\
				&$\theta=0.5$&65.95&\multicolumn{1}{c|}{8.95}&62.69&\multicolumn{1}{c|}{9.87}&62.50&9.05\\
				\midrule\specialrule{0.0em}{0pt}{0pt}
				\multirow{2}{*}{$\mathcal{D}_{ic}$}
				&$L1$&\bf 70.45 &\multicolumn{1}{c|}{5.22}&\bf 66.56&\multicolumn{1}{c|}{6.11}&63.31&8.24\\
				&$k=4$&69.01&\multicolumn{1}{c|}{3.52}&65.46&\multicolumn{1}{c|}{6.47}&66.04&5.51\\
				\midrule\specialrule{0.0em}{0pt}{0pt}
				\multicolumn{2}{l|}{MVCD}&69.70&\multicolumn{1}{c|}{\bf 3.41}&66.53&\multicolumn{1}{c|}{\bf 5.49}&\textbf{66.08}&\multicolumn{1}{c}{\bf 5.47}\\
				\bottomrule
		\end{tabular}}
	\end{center}
	
	\label{table:alter}
\end{table}

\begin{table}
	\caption{Training time ($second$).}
	\begin{center}
		\resizebox{0.75\linewidth}{!}{
			\begin{tabular}{l|ccc}
				\toprule
				Method&1&5&10\\
			    \midrule\specialrule{0.0em}{0pt}{0pt}
				MVCD&2254.52&	8227.90&	24001.84\\
				From Scratch&\multicolumn{3}{c}{36006.98}\\
				\bottomrule
		\end{tabular}}
	\end{center}
	
	\label{table:speed}
\end{table}

\subsection{Experiment Setup}
\textbf{Datasets.} The proposed method is evaluated on two benchmark datasets Pascal VOC 2007 and Microsoft COCO. VOC2007 has 20 object classes, and we use the trainval subset for training and the test subset for evaluation. COCO has 80K images in the training set and 40K images in the validation set for 80 object classes, and the minival (the first 5K images from the validation set) split is used for evaluation. There are two schemes to add new classes for evaluation: addition at once and sequential addition. 
	
	\noindent\textbf{Evaluation Metrics.} The compared methods are fine-tuning and some recent related works~\cite{shmelkov2017incremental}~\cite{chen2019new}~\cite{hao2019take}~\cite{hao2019end}~\cite{li2019rilod}. We reproduce the distillation methods and evaluate their performance under the same settings as our proposed method. We also design a baseline (Plain L1) that directly minimizes the L1 loss between the activations in the features of the old model and the incremental model. The basic object detector is Faster R-CNN for all methods. 
	We use both mean average precision (mAP) at 0.5 IoU threshold and the proposed ``SPmAP'' to measure the performance.
	\noindent\textbf{Implementation Details.} The old model is trained for 20 epochs, and the initial learning rate is set to 0.001 ($lr=0.001$), and decays every 5 epochs with $gamma=0.1$. The momentum is set to 0.9. The incremental model is trained for 10 epochs with $lr=0.0001$ and decays to 0.00001 after 5 epochs. The confidence and IoU threshold for NMS are set to 0.5 and 0.3 respectively. The thresholds in Section~\ref{sec:pc} are set to $\theta_{high}=0.8$ and $\theta_{low}=0.1$, and $k$ in Section~\ref{sec:ic} is set to 2. ResNet-50~\cite{he2016deep} is used as the backbone. We conduct all experiments on a single NVIDIA GeForce RTX 2080 Ti.
\subsection{Addition of Classes at Once}

In the first experiment, we evaluate the performance on adding new classes at once. We take 19, 15 and 10 classes from VOC2007 sorted in alphabetical order as the old classes, and the remaining 1, 5, 10 classes are the corresponding new classes as described in~\cite{shmelkov2017incremental}. For COCO, we take the first 40 classes as the old classes and the remaining 40 classes as the new classes. In these settings, if the image contains the categories to be detected, it will be selected for training or testing, so there is an overlap between the old data and the new data. However, the annotations of old classes in the new data are not available.

Table~\ref{table:all-detail} lists the per-category average precision on VOC2007 test subset. 
Old($\cdot$) represents the model trained on the old data, and Up-bound($\cdot$) represents the model trained on all data of both old and new classes. On the first setting, the mAP of fine-tuning gets only 26.2\%, which has caused severely catastrophic forgetting. Different from the original fine-tuning, which randomly initializes the classification layer for a new task, in order to preserve the learned knowledge, we initialize the parameters in the classification layer and the regression layer of the incremental model with those of the old model learned from the old classes. 
However, the performance of fine-tuning still degrades a lot. As can be seen, when only add one new class (``tv monitor"), the mAP and SPmAP of MVCD can reach 69.7\% and 3.4\% respectively, outperforming other L1/L2-distillation-based incremental object detection methods by a large margin. The mAP of MVCD exceeds the suboptimal Plain L1 about 0.8\%. It represents our method can better balance stability and plasticity.

For the second setting, we take 15 classes as the old classes, and the remaining 5 classes are added at once. MVCD also performs well compared with other methods. The mAP increases by about 2.0\% compared with Plain L1. With the increasing number of new classes, the mAP of fine-tuning is improved, however, the phenomenon of catastrophic forgetting is not mitigated. On the third setting, when 10 classes are added at once, the mAP of MVCD gets 66.1\%, outperforming the suboptimal Plain L1 about 0.5\%. Similarly, MVCD achieves the best SPmAP compared with other methods.


We also evaluate the performance on adding more classes as shown in Table~\ref{table:coco}. We take 40 classes from COCO training dataset as the old classes and the remaining 40 classes as the new classes. Both mAP and SPmAP of MVCD outperform Plain L1 and exceeds fine-tuning by a large margin.

The above results demonstrate that MVCD can effectively mitigate catastrophic forgetting on the setting of addition at once. The comparisons with other methods with the metric ``SPmAP'' also verify the superiority of MVCD on maintaining the stability and plasticity of the incremental model. 
As can be seen, the designed ``Plain L1'' achieves comparable performance with other first-order-distillation-based methods. Therefore, in the following experiments, we use ``Plain L1'' as the baseline for comparison.

\subsection{Sequential Addition of Multiple Classes}

In this experiment, we evaluate the performance of our method by adding classes sequentially for incremental learning. 
For the first setting, we also take 15 and 10 classes from VOC2007 sorted in alphabetical order as old classes, and the remaining 5 and 10 classes are as new classes. Table~\ref{table:10-sqe} lists the mAP(\%) when adding 5 and 10 classes sequentially.

As can be seen, MVCD in the setting of sequential addition of 5 classes outperforms Plain L1 in all incremental learning steps, and it can reach 51.89\% after the $5^{th}$ incremental learning step. The average improvements over all steps is 1.6\%, and the max difference can reach 3.25\% in the $2^{th}$ step.
We also evaluate the performance on adding 10 new classes sequentially with ten-step and five-step incremental learning respectively. In the ten-step learning, we add one new class at a time step, and in the five-step learning, we add two new classes at a time step.
As shown in the ten-step setting result, the proposed MVCD has consistent improvements in all learning steps. After the $6^{th}$ learning step, MVCD still exceeds Plain L1 6.78\% (40.05\% vs. 33.27\%). Due to the small number of samples in some categories, the performance is decreased slightly. However, MVCD is still better than Plain L1, which demonstrates the effectiveness of multi-view correlation distillation. In the five-step setting, the mAP of MVCD can still reach 48.23\% after the $5^{th}$ incremental learning step, and it outperforms Plain L1 by a large margin in all learning steps. 
These experiments demonstrate that the proposed MVCD can mitigate catastrophic forgetting better than the first-order distillation even after many incremental learning steps.


We also split the training set of VOC2007 and COCO into four groups: A, B, C and D as described in~\cite{hao2019end}. For fair comparisons, we also use ResNet-50~\cite{he2016deep} in this experiment. For each group, images that only contain the objects of classes in this group are selected, which means that there are no overlaps in these four groups. 
The results are shown in Table~\ref{table:4-group}, the performance of MVCD is better than Plain L1 in all incremental learning steps. On VOC2007, MVCD improves about 6.01\% compared with Plain L1 after the last learning step. On COCO, MVCD is consistently better than Plain L1 in all learning steps. 


\subsection{Ablation Study}

As listed in Table~\ref{table:ablation}, the proposed multi-view correlation distillation losses $\mathcal{D}_{cc}$, $\mathcal{D}_{pc}$ and $\mathcal{D}_{ic}$ on three settings are evaluated separately. The baseline only uses the distillation on the final classification and regression layers as shown in the first row in Table~\ref{table:ablation}. ``$+$" represents the increased mAP(\%) compared with the baseline.  

Firstly, these three distillation losses are evaluated individually. As can be seen, the accuracy increases by about 0.85\% on average by using $\mathcal{D}_{cc}$. 
The average increments of 1.85\% and 1.99\% are obtained when $\mathcal{D}_{pc}$ and $\mathcal{D}_{ic}$ are individually utilized. It verifies that these three correlation distillation losses are all useful for incremental object detection. The performance on SPmAP also show the effectiveness of these losses on preserving stability and plasticity. Then, we test different combinations of arbitrarily two losses, and the results on mAP show that the performances of these combinations are a little decreased compared with the combination of three correlation distillation losses.

The alternative choices of hyper-parameters $\theta$, $k$ and distillation ways on channel-wise, point-wise and instance-wise features are also tested as shown in Table~\ref{table:alter}. For MVCD in the last row, we use the settings as described in implementation details. For point-wise correlation distillation, we replace the high and low thresholds with a single threshold $\theta=0.5$ to divide the point-wise feature vectors into the vectors with high responses and low responses. Compared with our final setting $\theta_{high}=0.8$ and $\theta_{low}=0.1$, the performance decreases a lot, which verifies that only preserving the correlation between the most discriminative point-wise features can maintain the stability and plasticity of the incremental model better. For the instance-wise correlation distillation, the instance-level feature is divided into $4\times4$ ($k=4$) patches, and the result shows $k=2$ is better than $k=4$. We also replace the channel-wise, point-wise and instance-wise correlation distillation losses with L1 loss to minimize the distance between the selected features. The performance on SPmAP is worse than preserving the correlations, which demonstrates correlation distillation is more appropriate to get a tradeoff between stability and plasticity for incremental object detection.  

We also compare the training time of the proposed incremental learning method with training the detector from scratch using the similar GPU memory as listed in Table~\ref{table:speed}. When adding a few new classes, the proposed incremental object detection method has absolute superiority in training time with just a minor accuracy loss.
\section{Conclusion}
In this paper, we propose a novel multi-view correlation distillation based incremental object detection method, which transfers the correlations from the channel-wise, point-wise and instance-wise views in the feature space of the two-stage object detector. The channel-wise and point-wise correlations are designed for image-level features, and the instance-wise correlation is designed for instance-level features, which can get a good trade-off between the stability and the plasticity of the incremental model. Experimental results on VOC2007 and COCO with the new metric ``SPmAP'' demonstrate the effectiveness of the proposed method on incrementally learning to detect objects of new classes without severely forgetting originally learned knowledge.



\bibliographystyle{ACM-Reference-Format}
\bibliography{sample-base}


\begin{thebibliography}{38}


\ifx \showCODEN    \undefined \def \showCODEN     #1{\unskip}     \fi
\ifx \showDOI      \undefined \def \showDOI       #1{#1}\fi
\ifx \showISBNx    \undefined \def \showISBNx     #1{\unskip}     \fi
\ifx \showISBNxiii \undefined \def \showISBNxiii  #1{\unskip}     \fi
\ifx \showISSN     \undefined \def \showISSN      #1{\unskip}     \fi
\ifx \showLCCN     \undefined \def \showLCCN      #1{\unskip}     \fi
\ifx \shownote     \undefined \def \shownote      #1{#1}          \fi
\ifx \showarticletitle \undefined \def \showarticletitle #1{#1}   \fi
\ifx \showURL      \undefined \def \showURL       {\relax}        \fi
\providecommand\bibfield[2]{#2}
\providecommand\bibinfo[2]{#2}
\providecommand\natexlab[1]{#1}
\providecommand\showeprint[2][]{arXiv:#2}

\bibitem[\protect\citeauthoryear{Aljundi, Babiloni, Elhoseiny, Rohrbach, and
  Tuytelaars}{Aljundi et~al\mbox{.}}{2018}]%
        {aljundi2018memory}
\bibfield{author}{\bibinfo{person}{Rahaf Aljundi}, \bibinfo{person}{Francesca
  Babiloni}, \bibinfo{person}{Mohamed Elhoseiny}, \bibinfo{person}{Marcus
  Rohrbach}, {and} \bibinfo{person}{Tinne Tuytelaars}.}
  \bibinfo{year}{2018}\natexlab{}.
\newblock \showarticletitle{Memory aware synapses: Learning what (not) to
  forget}. In \bibinfo{booktitle}{\emph{Proceedings of the European Conference
  on Computer Vision}}. \bibinfo{pages}{139--154}.
\newblock


\bibitem[\protect\citeauthoryear{Aljundi, Chakravarty, and Tuytelaars}{Aljundi
  et~al\mbox{.}}{2017}]%
        {aljundi2017expert}
\bibfield{author}{\bibinfo{person}{Rahaf Aljundi}, \bibinfo{person}{Punarjay
  Chakravarty}, {and} \bibinfo{person}{Tinne Tuytelaars}.}
  \bibinfo{year}{2017}\natexlab{}.
\newblock \showarticletitle{Expert gate: Lifelong learning with a network of
  experts}. In \bibinfo{booktitle}{\emph{Proceedings of the IEEE Conference on
  Computer Vision and Pattern Recognition}}. \bibinfo{pages}{3366--3375}.
\newblock


\bibitem[\protect\citeauthoryear{Arbel{\'a}ez, Pont-Tuset, Barron, Marques, and
  Malik}{Arbel{\'a}ez et~al\mbox{.}}{2014}]%
        {arbelaez2014multiscale}
\bibfield{author}{\bibinfo{person}{Pablo Arbel{\'a}ez}, \bibinfo{person}{Jordi
  Pont-Tuset}, \bibinfo{person}{Jonathan~T Barron}, \bibinfo{person}{Ferran
  Marques}, {and} \bibinfo{person}{Jitendra Malik}.}
  \bibinfo{year}{2014}\natexlab{}.
\newblock \showarticletitle{Multiscale combinatorial grouping}. In
  \bibinfo{booktitle}{\emph{Proceedings of the IEEE Conference on Computer
  Vision and Pattern Recognition}}. \bibinfo{pages}{328--335}.
\newblock


\bibitem[\protect\citeauthoryear{Blanchard, Kinnison, RichardWebster, Bashivan,
  and Scheirer}{Blanchard et~al\mbox{.}}{2019}]%
        {blanchard2019neurobiological}
\bibfield{author}{\bibinfo{person}{Nathaniel Blanchard},
  \bibinfo{person}{Jeffery Kinnison}, \bibinfo{person}{Brandon RichardWebster},
  \bibinfo{person}{Pouya Bashivan}, {and} \bibinfo{person}{Walter~J Scheirer}.}
  \bibinfo{year}{2019}\natexlab{}.
\newblock \showarticletitle{A neurobiological evaluation metric for neural
  network model search}. In \bibinfo{booktitle}{\emph{Proceedings of the IEEE
  Conference on Computer Vision and Pattern Recognition}}.
  \bibinfo{pages}{5404--5413}.
\newblock


\bibitem[\protect\citeauthoryear{Chen, Yu, and Chen}{Chen
  et~al\mbox{.}}{2019}]%
        {chen2019new}
\bibfield{author}{\bibinfo{person}{Li Chen}, \bibinfo{person}{Chunyan Yu},
  {and} \bibinfo{person}{Lvcai Chen}.} \bibinfo{year}{2019}\natexlab{}.
\newblock \showarticletitle{A new knowledge distillation for incremental object
  detection}. In \bibinfo{booktitle}{\emph{2019 International Joint Conference
  on Neural Networks}}. IEEE, \bibinfo{pages}{1--7}.
\newblock


\bibitem[\protect\citeauthoryear{Everingham, Van~Gool, Williams, Winn, and
  Zisserman}{Everingham et~al\mbox{.}}{2010}]%
        {everingham2010pascal}
\bibfield{author}{\bibinfo{person}{Mark Everingham}, \bibinfo{person}{Luc
  Van~Gool}, \bibinfo{person}{Christopher~KI Williams}, \bibinfo{person}{John
  Winn}, {and} \bibinfo{person}{Andrew Zisserman}.}
  \bibinfo{year}{2010}\natexlab{}.
\newblock \showarticletitle{The {PASCAL} visual object classes ({VOC})
  challenge}.
\newblock \bibinfo{journal}{\emph{International Journal of Computer Vision}}
  \bibinfo{volume}{88}, \bibinfo{number}{2} (\bibinfo{year}{2010}),
  \bibinfo{pages}{303--338}.
\newblock


\bibitem[\protect\citeauthoryear{French}{French}{1999}]%
        {french1999catastrophic}
\bibfield{author}{\bibinfo{person}{Robert~M French}.}
  \bibinfo{year}{1999}\natexlab{}.
\newblock \showarticletitle{Catastrophic forgetting in connectionist networks}.
\newblock \bibinfo{journal}{\emph{Trends in Cognitive Sciences}}
  \bibinfo{volume}{3}, \bibinfo{number}{4} (\bibinfo{year}{1999}),
  \bibinfo{pages}{128--135}.
\newblock


\bibitem[\protect\citeauthoryear{Girshick}{Girshick}{2015}]%
        {girshick2015fast}
\bibfield{author}{\bibinfo{person}{Ross Girshick}.}
  \bibinfo{year}{2015}\natexlab{}.
\newblock \showarticletitle{Fast {R-CNN}}. In
  \bibinfo{booktitle}{\emph{Proceedings of the IEEE International Conference on
  Computer Vision}}. \bibinfo{pages}{1440--1448}.
\newblock


\bibitem[\protect\citeauthoryear{Goodfellow, Mirza, Xiao, Courville, and
  Bengio}{Goodfellow et~al\mbox{.}}{2013}]%
        {goodfellow2013empirical}
\bibfield{author}{\bibinfo{person}{Ian~J Goodfellow}, \bibinfo{person}{Mehdi
  Mirza}, \bibinfo{person}{Da Xiao}, \bibinfo{person}{Aaron Courville}, {and}
  \bibinfo{person}{Yoshua Bengio}.} \bibinfo{year}{2013}\natexlab{}.
\newblock \showarticletitle{An empirical investigation of catastrophic
  forgetting in gradient-based neural networks}.
\newblock \bibinfo{journal}{\emph{Computer Science}} \bibinfo{volume}{84},
  \bibinfo{number}{12} (\bibinfo{year}{2013}), \bibinfo{pages}{1387--91}.
\newblock


\bibitem[\protect\citeauthoryear{Hao, Fu, and Jiang}{Hao
  et~al\mbox{.}}{2019a}]%
        {hao2019take}
\bibfield{author}{\bibinfo{person}{Yu Hao}, \bibinfo{person}{Yanwei Fu}, {and}
  \bibinfo{person}{Yu-Gang Jiang}.} \bibinfo{year}{2019}\natexlab{a}.
\newblock \showarticletitle{Take goods from shelves: A dataset for
  class-incremental object detection}. In \bibinfo{booktitle}{\emph{Proceedings
  of the 2019 on International Conference on Multimedia Retrieval}}.
  \bibinfo{pages}{271--278}.
\newblock


\bibitem[\protect\citeauthoryear{Hao, Fu, Jiang, and Tian}{Hao
  et~al\mbox{.}}{2019b}]%
        {hao2019end}
\bibfield{author}{\bibinfo{person}{Yu Hao}, \bibinfo{person}{Yanwei Fu},
  \bibinfo{person}{Yu-Gang Jiang}, {and} \bibinfo{person}{Qi Tian}.}
  \bibinfo{year}{2019}\natexlab{b}.
\newblock \showarticletitle{An end-to-end architecture for class-incremental
  object detection with knowledge distillation}. In
  \bibinfo{booktitle}{\emph{2019 IEEE International Conference on Multimedia
  and Expo}}. IEEE, \bibinfo{pages}{1--6}.
\newblock


\bibitem[\protect\citeauthoryear{He, Zhang, Ren, and Sun}{He
  et~al\mbox{.}}{2016}]%
        {he2016deep}
\bibfield{author}{\bibinfo{person}{Kaiming He}, \bibinfo{person}{Xiangyu
  Zhang}, \bibinfo{person}{Shaoqing Ren}, {and} \bibinfo{person}{Jian Sun}.}
  \bibinfo{year}{2016}\natexlab{}.
\newblock \showarticletitle{Deep residual learning for image recognition}. In
  \bibinfo{booktitle}{\emph{Proceedings of the IEEE Conference on Computer
  Vision and Pattern Recognition}}. \bibinfo{pages}{770--778}.
\newblock


\bibitem[\protect\citeauthoryear{Hinton, Vinyals, and Dean}{Hinton
  et~al\mbox{.}}{2015}]%
        {hinton2015distilling}
\bibfield{author}{\bibinfo{person}{Geoffrey Hinton}, \bibinfo{person}{Oriol
  Vinyals}, {and} \bibinfo{person}{Jeff Dean}.}
  \bibinfo{year}{2015}\natexlab{}.
\newblock \showarticletitle{Distilling the knowledge in a neural network}.
\newblock \bibinfo{journal}{\emph{Computer Science}} \bibinfo{volume}{14},
  \bibinfo{number}{7} (\bibinfo{year}{2015}), \bibinfo{pages}{38--39}.
\newblock


\bibitem[\protect\citeauthoryear{Hou, Pan, Loy, Wang, and Lin}{Hou
  et~al\mbox{.}}{2019}]%
        {hou2019learning}
\bibfield{author}{\bibinfo{person}{Saihui Hou}, \bibinfo{person}{Xinyu Pan},
  \bibinfo{person}{Chen~Change Loy}, \bibinfo{person}{Zilei Wang}, {and}
  \bibinfo{person}{Dahua Lin}.} \bibinfo{year}{2019}\natexlab{}.
\newblock \showarticletitle{Learning a unified classifier incrementally via
  rebalancing}. In \bibinfo{booktitle}{\emph{Proceedings of the IEEE Conference
  on Computer Vision and Pattern Recognition}}. \bibinfo{pages}{831--839}.
\newblock


\bibitem[\protect\citeauthoryear{Hu, Shen, and Sun}{Hu et~al\mbox{.}}{2018}]%
        {hu2018squeeze}
\bibfield{author}{\bibinfo{person}{Jie Hu}, \bibinfo{person}{Li Shen}, {and}
  \bibinfo{person}{Gang Sun}.} \bibinfo{year}{2018}\natexlab{}.
\newblock \showarticletitle{Squeeze-and-excitation networks}. In
  \bibinfo{booktitle}{\emph{Proceedings of the IEEE Conference on Computer
  Vision and Pattern Recognition}}. \bibinfo{pages}{7132--7141}.
\newblock


\bibitem[\protect\citeauthoryear{Jung, Ju, Jung, and Kim}{Jung
  et~al\mbox{.}}{2018}]%
        {jung2018less}
\bibfield{author}{\bibinfo{person}{Heechul Jung}, \bibinfo{person}{Jeongwoo
  Ju}, \bibinfo{person}{Minju Jung}, {and} \bibinfo{person}{Junmo Kim}.}
  \bibinfo{year}{2018}\natexlab{}.
\newblock \showarticletitle{Less-forgetful learning for domain expansion in
  deep neural networks}. In \bibinfo{booktitle}{\emph{Thirty-Second AAAI
  Conference on Artificial Intelligence}}. \bibinfo{pages}{3358--3365}.
\newblock


\bibitem[\protect\citeauthoryear{Kirkpatrick, Pascanu, Rabinowitz, Veness,
  Desjardins, Rusu, Milan, Quan, Ramalho, Grabska-Barwinska,
  et~al\mbox{.}}{Kirkpatrick et~al\mbox{.}}{2017}]%
        {kirkpatrick2017overcoming}
\bibfield{author}{\bibinfo{person}{James Kirkpatrick}, \bibinfo{person}{Razvan
  Pascanu}, \bibinfo{person}{Neil Rabinowitz}, \bibinfo{person}{Joel Veness},
  \bibinfo{person}{Guillaume Desjardins}, \bibinfo{person}{Andrei~A Rusu},
  \bibinfo{person}{Kieran Milan}, \bibinfo{person}{John Quan},
  \bibinfo{person}{Tiago Ramalho}, \bibinfo{person}{Agnieszka
  Grabska-Barwinska}, {et~al\mbox{.}}} \bibinfo{year}{2017}\natexlab{}.
\newblock \showarticletitle{Overcoming catastrophic forgetting in neural
  networks}.
\newblock \bibinfo{journal}{\emph{Proceedings of the National Academy of
  Sciences}} \bibinfo{volume}{114}, \bibinfo{number}{13}
  (\bibinfo{year}{2017}), \bibinfo{pages}{3521--3526}.
\newblock


\bibitem[\protect\citeauthoryear{Kriegeskorte, Mur, and
  Bandettini}{Kriegeskorte et~al\mbox{.}}{2008}]%
        {kriegeskorte2008representational}
\bibfield{author}{\bibinfo{person}{Nikolaus Kriegeskorte},
  \bibinfo{person}{Marieke Mur}, {and} \bibinfo{person}{Peter~A Bandettini}.}
  \bibinfo{year}{2008}\natexlab{}.
\newblock \showarticletitle{Representational similarity analysis-connecting the
  branches of systems neuroscience}.
\newblock \bibinfo{journal}{\emph{Frontiers in Systems Neuroscience}}
  \bibinfo{volume}{2} (\bibinfo{year}{2008}), \bibinfo{pages}{4}.
\newblock


\bibitem[\protect\citeauthoryear{Li, Tasci, Ghosh, Zhu, Zhang, and Heck}{Li
  et~al\mbox{.}}{2019}]%
        {li2019rilod}
\bibfield{author}{\bibinfo{person}{Dawei Li}, \bibinfo{person}{Serafettin
  Tasci}, \bibinfo{person}{Shalini Ghosh}, \bibinfo{person}{Jingwen Zhu},
  \bibinfo{person}{Junting Zhang}, {and} \bibinfo{person}{Larry Heck}.}
  \bibinfo{year}{2019}\natexlab{}.
\newblock \showarticletitle{{RILOD}: Near real-time incremental learning for
  object detection at the edge}. In \bibinfo{booktitle}{\emph{Proceedings of
  the 4th ACM/IEEE Symposium on Edge Computing}}. \bibinfo{pages}{113--126}.
\newblock


\bibitem[\protect\citeauthoryear{Li, Wu, Fang, Liao, Wang, and Qian}{Li
  et~al\mbox{.}}{2020}]%
        {li2020local}
\bibfield{author}{\bibinfo{person}{Xiaojie Li}, \bibinfo{person}{Jianlong Wu},
  \bibinfo{person}{Hongyu Fang}, \bibinfo{person}{Yue Liao},
  \bibinfo{person}{Fei Wang}, {and} \bibinfo{person}{Chen Qian}.}
  \bibinfo{year}{2020}\natexlab{}.
\newblock \showarticletitle{Local correlation consistency for knowledge
  distillation}. In \bibinfo{booktitle}{\emph{European Conference on Computer
  Vision}}. Springer, \bibinfo{pages}{18--33}.
\newblock


\bibitem[\protect\citeauthoryear{Li and Hoiem}{Li and Hoiem}{2017}]%
        {li2017learning}
\bibfield{author}{\bibinfo{person}{Zhizhong Li} {and} \bibinfo{person}{Derek
  Hoiem}.} \bibinfo{year}{2017}\natexlab{}.
\newblock \showarticletitle{Learning without forgetting}.
\newblock \bibinfo{journal}{\emph{IEEE Transactions on Pattern Analysis and
  Machine Intelligence}} \bibinfo{volume}{40}, \bibinfo{number}{12}
  (\bibinfo{year}{2017}), \bibinfo{pages}{2935--2947}.
\newblock


\bibitem[\protect\citeauthoryear{Lin, Goyal, Girshick, He, and Doll{\'a}r}{Lin
  et~al\mbox{.}}{2017}]%
        {lin2017focal}
\bibfield{author}{\bibinfo{person}{Tsung-Yi Lin}, \bibinfo{person}{Priya
  Goyal}, \bibinfo{person}{Ross Girshick}, \bibinfo{person}{Kaiming He}, {and}
  \bibinfo{person}{Piotr Doll{\'a}r}.} \bibinfo{year}{2017}\natexlab{}.
\newblock \showarticletitle{Focal loss for dense object detection}. In
  \bibinfo{booktitle}{\emph{Proceedings of the IEEE International Conference on
  Computer Vision}}. \bibinfo{pages}{2980--2988}.
\newblock


\bibitem[\protect\citeauthoryear{Lin, Maire, Belongie, Hays, Perona, Ramanan,
  Doll{\'a}r, and Zitnick}{Lin et~al\mbox{.}}{2014}]%
        {lin2014microsoft}
\bibfield{author}{\bibinfo{person}{Tsung-Yi Lin}, \bibinfo{person}{Michael
  Maire}, \bibinfo{person}{Serge Belongie}, \bibinfo{person}{James Hays},
  \bibinfo{person}{Pietro Perona}, \bibinfo{person}{Deva Ramanan},
  \bibinfo{person}{Piotr Doll{\'a}r}, {and} \bibinfo{person}{C~Lawrence
  Zitnick}.} \bibinfo{year}{2014}\natexlab{}.
\newblock \showarticletitle{Microsoft {COCO}: Common objects in context}. In
  \bibinfo{booktitle}{\emph{Proceedings of the European Conference on Computer
  Vision}}. Springer, \bibinfo{pages}{740--755}.
\newblock


\bibitem[\protect\citeauthoryear{Liu, Cao, Li, Yuan, Hu, Li, and Duan}{Liu
  et~al\mbox{.}}{2019}]%
        {liu2019knowledge}
\bibfield{author}{\bibinfo{person}{Yufan Liu}, \bibinfo{person}{Jiajiong Cao},
  \bibinfo{person}{Bing Li}, \bibinfo{person}{Chunfeng Yuan},
  \bibinfo{person}{Weiming Hu}, \bibinfo{person}{Yangxi Li}, {and}
  \bibinfo{person}{Yunqiang Duan}.} \bibinfo{year}{2019}\natexlab{}.
\newblock \showarticletitle{Knowledge distillation via instance relationship
  graph}. In \bibinfo{booktitle}{\emph{Proceedings of the IEEE Conference on
  Computer Vision and Pattern Recognition}}. \bibinfo{pages}{7096--7104}.
\newblock


\bibitem[\protect\citeauthoryear{Matthews}{Matthews}{2001}]%
        {matthews2001short}
\bibfield{author}{\bibinfo{person}{Peter Matthews}.}
  \bibinfo{year}{2001}\natexlab{}.
\newblock \bibinfo{booktitle}{\emph{A short history of structural
  linguistics}}.
\newblock \bibinfo{publisher}{Cambridge University Press}.
\newblock


\bibitem[\protect\citeauthoryear{McCloskey and Cohen}{McCloskey and
  Cohen}{1989}]%
        {mccloskey1989catastrophic}
\bibfield{author}{\bibinfo{person}{Michael McCloskey} {and}
  \bibinfo{person}{Neal~J Cohen}.} \bibinfo{year}{1989}\natexlab{}.
\newblock \showarticletitle{Catastrophic interference in connectionist
  networks: The sequential learning problem}.
\newblock In \bibinfo{booktitle}{\emph{Psychology of Learning and Motivation}}.
  Vol.~\bibinfo{volume}{24}. \bibinfo{publisher}{Elsevier},
  \bibinfo{pages}{109--165}.
\newblock


\bibitem[\protect\citeauthoryear{Park, Kim, Lu, and Cho}{Park
  et~al\mbox{.}}{2019}]%
        {park2019relational}
\bibfield{author}{\bibinfo{person}{Wonpyo Park}, \bibinfo{person}{Dongju Kim},
  \bibinfo{person}{Yan Lu}, {and} \bibinfo{person}{Minsu Cho}.}
  \bibinfo{year}{2019}\natexlab{}.
\newblock \showarticletitle{Relational knowledge distillation}. In
  \bibinfo{booktitle}{\emph{Proceedings of the IEEE Conference on Computer
  Vision and Pattern Recognition}}. \bibinfo{pages}{3967--3976}.
\newblock


\bibitem[\protect\citeauthoryear{Perez-Rua, Zhu, Hospedales, and
  Xiang}{Perez-Rua et~al\mbox{.}}{2020}]%
        {perez2020incremental}
\bibfield{author}{\bibinfo{person}{Juan-Manuel Perez-Rua},
  \bibinfo{person}{Xiatian Zhu}, \bibinfo{person}{Timothy~M Hospedales}, {and}
  \bibinfo{person}{Tao Xiang}.} \bibinfo{year}{2020}\natexlab{}.
\newblock \showarticletitle{Incremental few-shot object detection}. In
  \bibinfo{booktitle}{\emph{Proceedings of the IEEE Conference on Computer
  Vision and Pattern Recognition}}. \bibinfo{pages}{13846--13855}.
\newblock


\bibitem[\protect\citeauthoryear{Rannen, Aljundi, Blaschko, and
  Tuytelaars}{Rannen et~al\mbox{.}}{2017}]%
        {rannen2017encoder}
\bibfield{author}{\bibinfo{person}{Amal Rannen}, \bibinfo{person}{Rahaf
  Aljundi}, \bibinfo{person}{Matthew~B Blaschko}, {and} \bibinfo{person}{Tinne
  Tuytelaars}.} \bibinfo{year}{2017}\natexlab{}.
\newblock \showarticletitle{Encoder based lifelong learning}. In
  \bibinfo{booktitle}{\emph{Proceedings of the IEEE International Conference on
  Computer Vision}}. \bibinfo{pages}{1320--1328}.
\newblock


\bibitem[\protect\citeauthoryear{Rebuffi, Kolesnikov, Sperl, and
  Lampert}{Rebuffi et~al\mbox{.}}{2017}]%
        {rebuffi2017icarl}
\bibfield{author}{\bibinfo{person}{Sylvestre-Alvise Rebuffi},
  \bibinfo{person}{Alexander Kolesnikov}, \bibinfo{person}{Georg Sperl}, {and}
  \bibinfo{person}{Christoph~H Lampert}.} \bibinfo{year}{2017}\natexlab{}.
\newblock \showarticletitle{{iCaRL}: Incremental classifier and representation
  learning}. In \bibinfo{booktitle}{\emph{Proceedings of the IEEE Conference on
  Computer Vision and Pattern Recognition}}. \bibinfo{pages}{2001--2010}.
\newblock


\bibitem[\protect\citeauthoryear{Ren, He, Girshick, and Sun}{Ren
  et~al\mbox{.}}{2015}]%
        {ren2015faster}
\bibfield{author}{\bibinfo{person}{Shaoqing Ren}, \bibinfo{person}{Kaiming He},
  \bibinfo{person}{Ross Girshick}, {and} \bibinfo{person}{Jian Sun}.}
  \bibinfo{year}{2015}\natexlab{}.
\newblock \showarticletitle{Faster {R-CNN}: Towards real-time object detection
  with region proposal networks}. In \bibinfo{booktitle}{\emph{Advances in
  Neural Information Processing Systems}}. \bibinfo{pages}{91--99}.
\newblock


\bibitem[\protect\citeauthoryear{Shmelkov, Schmid, and Alahari}{Shmelkov
  et~al\mbox{.}}{2017}]%
        {shmelkov2017incremental}
\bibfield{author}{\bibinfo{person}{Konstantin Shmelkov},
  \bibinfo{person}{Cordelia Schmid}, {and} \bibinfo{person}{Karteek Alahari}.}
  \bibinfo{year}{2017}\natexlab{}.
\newblock \showarticletitle{Incremental learning of object detectors without
  catastrophic forgetting}. In \bibinfo{booktitle}{\emph{Proceedings of the
  IEEE International Conference on Computer Vision}}.
  \bibinfo{pages}{3400--3409}.
\newblock


\bibitem[\protect\citeauthoryear{Sun, Cong, and Xu}{Sun et~al\mbox{.}}{2018a}]%
        {sun2018active}
\bibfield{author}{\bibinfo{person}{Gan Sun}, \bibinfo{person}{Yang Cong}, {and}
  \bibinfo{person}{Xiaowei Xu}.} \bibinfo{year}{2018}\natexlab{a}.
\newblock \showarticletitle{Active lifelong learning with ``watchdog"}. In
  \bibinfo{booktitle}{\emph{Thirty-Second AAAI Conference on Artificial
  Intelligence}}. \bibinfo{pages}{4107--4114}.
\newblock


\bibitem[\protect\citeauthoryear{Sun, Yang, Liu, Liu, Xu, and Yu}{Sun
  et~al\mbox{.}}{2018b}]%
        {sun2018lifelong}
\bibfield{author}{\bibinfo{person}{Gan Sun}, \bibinfo{person}{Cong Yang},
  \bibinfo{person}{Ji Liu}, \bibinfo{person}{Lianqing Liu},
  \bibinfo{person}{Xiaowei Xu}, {and} \bibinfo{person}{Haibin Yu}.}
  \bibinfo{year}{2018}\natexlab{b}.
\newblock \showarticletitle{Lifelong metric learning}.
\newblock \bibinfo{journal}{\emph{IEEE Transactions on Cybernetics}}
  \bibinfo{volume}{49}, \bibinfo{number}{8} (\bibinfo{year}{2018}),
  \bibinfo{pages}{3168--3179}.
\newblock


\bibitem[\protect\citeauthoryear{Yang, Wang, Chen, Liu, and Qiao}{Yang
  et~al\mbox{.}}{2020}]%
        {yang2020context}
\bibfield{author}{\bibinfo{person}{Ze Yang}, \bibinfo{person}{Yali Wang},
  \bibinfo{person}{Xianyu Chen}, \bibinfo{person}{Jianzhuang Liu}, {and}
  \bibinfo{person}{Yu Qiao}.} \bibinfo{year}{2020}\natexlab{}.
\newblock \showarticletitle{{Context-Transformer}: Tackling object confusion
  for few-shot detection}. In \bibinfo{booktitle}{\emph{Thirty-Fourth AAAI
  Conference on Artificial Intelligence}}. \bibinfo{pages}{12653--12660}.
\newblock


\bibitem[\protect\citeauthoryear{Zenke, Poole, and Ganguli}{Zenke
  et~al\mbox{.}}{2017}]%
        {zenke2017continual}
\bibfield{author}{\bibinfo{person}{Friedemann Zenke}, \bibinfo{person}{Ben
  Poole}, {and} \bibinfo{person}{Surya Ganguli}.}
  \bibinfo{year}{2017}\natexlab{}.
\newblock \showarticletitle{Continual learning through synaptic intelligence}.
  In \bibinfo{booktitle}{\emph{International Conference on Machine Learning}}.
  PMLR, \bibinfo{pages}{3987--3995}.
\newblock


\bibitem[\protect\citeauthoryear{Zhang, Zhang, Ghosh, Li, Tasci, Heck, Zhang,
  and Kuo}{Zhang et~al\mbox{.}}{2020}]%
        {zhang2020class}
\bibfield{author}{\bibinfo{person}{Junting Zhang}, \bibinfo{person}{Jie Zhang},
  \bibinfo{person}{Shalini Ghosh}, \bibinfo{person}{Dawei Li},
  \bibinfo{person}{Serafettin Tasci}, \bibinfo{person}{Larry Heck},
  \bibinfo{person}{Heming Zhang}, {and} \bibinfo{person}{C-C~Jay Kuo}.}
  \bibinfo{year}{2020}\natexlab{}.
\newblock \showarticletitle{Class-incremental learning via deep model
  consolidation}. In \bibinfo{booktitle}{\emph{The IEEE Winter Conference on
  Applications of Computer Vision}}. \bibinfo{pages}{1131--1140}.
\newblock


\bibitem[\protect\citeauthoryear{Zitnick and Doll{\'a}r}{Zitnick and
  Doll{\'a}r}{2014}]%
        {zitnick2014edge}
\bibfield{author}{\bibinfo{person}{C~Lawrence Zitnick} {and}
  \bibinfo{person}{Piotr Doll{\'a}r}.} \bibinfo{year}{2014}\natexlab{}.
\newblock \showarticletitle{Edge boxes: Locating object proposals from edges}.
  In \bibinfo{booktitle}{\emph{Proceedings of the European Conference on
  Computer Vision}}. Springer, \bibinfo{pages}{391--405}.
\newblock


\end{thebibliography}

%
%

\end{sloppypar}
\end{document}